\documentclass[11pt,a4paper]{article}
\usepackage[left=2cm, right=2cm, top=2.25cm, bottom=2.25cm,  headheight=12pt, a4paper]{geometry}%
\usepackage{authblk} 
\usepackage{lipsum}
\usepackage[utf8]{inputenc}
\usepackage[T1]{fontenc}
\usepackage{microtype,newtxtext,newtxmath} 
\usepackage{mathtools,gensymb,eurosym,url,soul} 
\usepackage{caption}
\usepackage{subcaption}
\usepackage{longtable,booktabs,array,multirow}
\usepackage[normalem]{ulem}
\usepackage[backend=biber, sorting=none, maxnames=50]{biblatex}
\usepackage[switch]{lineno}
\usepackage{acronym}
\usepackage{graphicx}
\usepackage{svg}
\usepackage{xcolor}
\usepackage{xr,xr-hyper} 
\usepackage{hyperref}
\usepackage[subpreambles=true]{standalone} 
\usepackage{algorithm}
\usepackage{algpseudocode}
\usepackage{tikz}
\usepackage{pgfplots}
\usepgfplotslibrary{groupplots}
\usetikzlibrary{patterns}

\hypersetup{colorlinks=true, breaklinks=true, 
  urlcolor=blue, linkcolor=blue,anchorcolor=blue,citecolor=blue,
  hypertexnames=true, final=true,
  pdfpagemode = UseNone, 
  pdfauthor = {},
  pdftitle = {},
  pdfsubject = {},
  pdfkeywords = {}
}

\graphicspath{{img/} {figures/} {./} {figures_SM/}}
\DeclareGraphicsExtensions{.pdf,.png,.svg,.jpg,.mps}

\DeclareSourcemap{
  \maps[datatype=bibtex]{
    \map[overwrite=true]{
      \step[fieldset=url, null]
      \step[fieldset=eprint, null]
    }
  }
}

\acrodef{ADC}[ADC]{Analog-to-Digital Converter}
\acrodef{ADEXP}[AdExp-IF]{Adaptive Exponential Integrate-and-Fire}
\acrodef{ADM}[ADM]{Asynchronous Delta Modulator}
\acrodef{AER}[AER]{Address-Event Representation}
\acrodef{AEX}[AEX]{AER EXtension board}
\acrodef{AE}[AE]{Address-Event}
\acrodef{AFE}[AFE]{Analog Front-End}
\acrodef{AFM}[AFM]{Atomic Force Microscope}
\acrodef{AGC}[AGC]{Automatic Gain Control}
\acrodef{AI}[AI]{Artificial Intelligence}
\acrodef{AMDA}[AMDA]{AER Motherboard with D/A converters}
\acrodef{AMPA}[AMPA]{$\alpha$-Amino-3-hydroxy-5-methyl-4-isoxazolepropionic Acid}
\acrodef{ANN}[ANN]{Artificial Neural Network}
\acrodef{API}[API]{Application Programming Interface}
\acrodef{APMOM}[APMOM]{Alternate Polarity Metal On Metal}
\acrodef{ARM}[ARM]{Advanced RISC Machine}
\acrodef{ASIC}[ASIC]{Application Specific Integrated Circuit}
\acrodef{BCM}[BMC]{Bienenstock-Cooper-Munro}
\acrodef{BD}[BD]{Bundled Data}
\acrodef{BEOL}[BEOL]{Back-end of Line}
\acrodef{BG}[BG]{Bias Generator}
\acrodef{BMI}[BMI]{Brain-Machince Interface}
\acrodef{BTB}[BTB]{Band-to-Band tunnelling}
\acrodef{BTSP}[BTSP]{Behavioral Time Scale Synaptic Plasticity}
\acrodef{CAD}[CAD]{Computer Aided Design}
\acrodef{CAM}[CAM]{Content Addressable Memory}
\acrodef{CAVIAR}[CAVIAR]{Convolution AER Vision Architecture for Real-Time}
\acrodef{CA}[CA]{Cortical Automaton}
\acrodef{CCN}[CCN]{Cooperative and Competitive Network}
\acrodef{CDR}[CDR]{Clock-Data Recovery}
\acrodef{CFC}[CFC]{Current to Frequency Converter}
\acrodef{CHP}[CHP]{Communicating Hardware Processes}
\acrodef{CMIM}[CMIM]{Metal-Insulator-Metal Capacitor}
\acrodef{CML}[CML]{Current Mode Logic}
\acrodef{CMOL}[CMOL]{Hybrid CMOS nanoelectronic circuits}
\acrodef{CMOS}[CMOS]{Complementary Metal-Oxide-Semiconductor}
\acrodef{CNN}[CNN]{Convolutional Neural Network}
\acrodef{CNS}[CNS]{central Nervous System}
\acrodef{COTS}[COTS]{Commercial Off-The-Shelf}
\acrodef{CPG}[CPG]{Central Pattern Generator}
\acrodef{CPLD}[CPLD]{Complex Programmable Logic Device}
\acrodef{CPU}[CPU]{Central Processing Unit}
\acrodef{CSM}[CSM]{Cortical State Machine}
\acrodef{CSP}[CSP]{Constraint Satisfaction Problem}
\acrodef{CTXCTL}[CTXCTL]{CortexControl}
\acrodef{CV}[CV]{Coefficient of Variation}
\acrodef{DAC}[DAC]{Digital to Analog Converter}
\acrodef{DAS}[DAS]{Dynamic Auditory Sensor}
\acrodef{DAVIS}[DAVIS]{Dynamic and Active Pixel Vision Sensor}
\acrodef{DBN}[DBN]{Deep Belief Network}
\acrodef{DBS}[DBS]{Deep Brain Stimulation}
\acrodef{DFA}[DFA]{Deterministic Finite Automaton}
\acrodef{DIBL}[DIBL]{Drain-Induced Barrier-Lowering}
\acrodef{DI}[DI]{Delay Insensitive}
\acrodef{DMA}[DMA]{Direct Memory Access}
\acrodef{DNF}[DNF]{Dynamic Neural Field}
\acrodef{DNN}[DNN]{Deep Neural Network}
\acrodef{DOF}[DOF]{Degrees of Freedom}
\acrodef{DPE}[DPE]{Dynamic Parameter Estimation}
\acrodef{DPI}[DPI]{Differential Pair Integrator}
\acrodef{DRAM}[DRAM]{Dynamic Random Access Memory}
\acrodef{DRRZ}[DR-RZ]{Dual-Rail Return-to-Zero}
\acrodef{DR}[DR]{Dual Rail}
\acrodef{DSP}[DSP]{Digital Signal Processor}
\acrodef{DVS}[DVS]{Dynamic Vision Sensor}
\acrodef{DYNAP}[DYNAP]{Dynamic Neuromorphic Asynchronous Processor}
\acrodef{EBL}[EBL]{Electron Beam Lithography}
\acrodef{ECG}[ECG]{Electrocardiography}
\acrodef{ECoG}[ECoG]{Electrocorticography}
\acrodef{EDVAC}[EDVAC]{Electronic Discrete Variable Automatic Computer}
\acrodef{EEG}[EEG]{Electroencephalography}
\acrodef{EI}[EI]{Excitatory-Inhibitory}
\acrodef{EIN}[EIN]{Excitatory-Inhibitory Network}
\acrodef{EMG}[EMG]{Electromyography}
\acrodef{EM}[EM]{Expectation Maximization}
\acrodef{EOG}[EOG]{Electrooculogram}
\acrodef{EPSC}[EPSC]{Excitatory Post-Synaptic Current}
\acrodef{EPSP}[EPSP]{Excitatory Post-Synaptic Potential}
\acrodef{EZ}[EZ]{Epileptogenic Zone}
\acrodef{FDSOI}[FDSOI]{Fully-Depleted Silicon on Insulator}
\acrodef{FET}[FET]{Field-Effect Transistor}
\acrodef{FFT}[FFT]{Fast Fourier Transform}
\acrodef{FI}[F-I]{Frequency--Current}
\acrodef{FMA}[FMA]{Floating Microelectrode Array}
\acrodef{FNN}[FNN]{Feed-forward Neural Network}
\acrodef{FPGA}[FPGA]{Field Programmable Gate Array}
\acrodef{FR}[FR]{Fast Ripple}
\acrodef{FSA}[FSA]{Finite State Automaton}
\acrodef{FSM}[FSM]{Finite State Machine}
\acrodef{GABA}[GABA]{$\gamma$-Aminobutanoic Acid}
\acrodef{GIDL}[GIDL]{Gate-Induced Drain Leakage}
\acrodef{GOPS}[GOPS]{Giga-Operations per Second}
\acrodef{GPIO}[GPIO]{General Purpose I/O}
\acrodef{GPU}[GPU]{Graphical Processing Unit}
\acrodef{GT}[GT]{Ground Truth}
\acrodef{GUI}[GUI]{Graphical User Interface}
\acrodef{HAL}[HAL]{Hardware Abstraction Layer}
\acrodef{HFO}[HFO]{High Frequency Oscillation}
\acrodef{HH}[H\&H]{Hodgkin \& Huxley}
\acrodef{HMM}[HMM]{Hidden Markov Model}
\acrodef{HRS}[HRS]{High-Resistive State}
\acrodef{HR}[HR]{Heart Rate}
\acrodef{HSE}[HSE]{Handshaking Expansion}
\acrodef{HW}[HW]{Hardware}
\acrodef{ICT}[ICT]{Information and Communication Technology}
\acrodef{IC}[IC]{Integrated Circuit}
\acrodef{IF2DWTA}[IF2DWTA]{Integrate \& Fire 2-Dimensional WTA}
\acrodef{IFSLWTA}[IFSLWTA]{Integrate \& Fire Stop Learning WTA}
\acrodef{IF}[I\&F]{Integrate-and-Fire}
\acrodef{IMU}[IMU]{Inertial Measurement Unit}
\acrodef{INCF}[INCF]{International Neuroinformatics Coordinating Facility}
\acrodef{INI}[INI]{Institute of Neuroinformatics}
\acrodef{IO}[I/O]{Input/Output}
\acrodef{IPSC}[IPSC]{Inhibitory Post-Synaptic Current}
\acrodef{IPSP}[IPSP]{Inhibitory Post-Synaptic Potential}
\acrodef{IP}[IP]{Intellectual Property}
\acrodef{ISI}[ISI]{Inter-Spike Interval}
\acrodef{IoT}[IoT]{Internet of Things}
\acrodef{JFLAP}[JFLAP]{Java - Formal Languages and Automata Package}
\acrodef{LEDR}[LEDR]{Level-Encoded Dual-Rail}
\acrodef{LFP}[LFP]{Local Field Potential}
\acrodef{LIFE}[LIFE]{Longitudinal Intrafascicular Electrodes}
\acrodef{LIF}[LI\&F]{Leaky Integrate-and-Fire}
\acrodef{LLC}[LLC]{Low Leakage Cell}
\acrodef{LNA}[LNA]{Low-Noise Amplifier}
\acrodef{LPF}[LPF]{Low Pass Filter}
\acrodef{LRS}[LRS]{Low-Resistive State}
\acrodef{LSM}[LSM]{Liquid State Machine}
\acrodef{LTD}[LTD]{Long Term Depression}
\acrodef{LTI}[LTI]{Linear Time-Invariant}
\acrodef{LTP}[LTP]{Long Term Potentiation}
\acrodef{LTU}[LTU]{Linear Threshold Unit}
\acrodef{LUT}[LUT]{Look-Up Table}
\acrodef{LVDS}[LVDS]{Low Voltage Differential Signaling}
\acrodef{MCMC}[MCMC]{Markov-Chain Monte Carlo}
\acrodef{MEA}[MEA]{Multielectrode Arrays}
\acrodef{MEMS}[MEMS]{Micro Electro Mechanical System}
\acrodef{MFR}[MFR]{Mean Firing Rate}
\acrodef{MIM}[MIM]{Metal Insulator Metal}
\acrodef{MLP}[MLP]{Multilayer Perceptron}
\acrodef{ML}[ML]{Machine Learning}
\acrodef{MOSCAP}[MOSCAP]{Metal Oxide Semiconductor Capacitor}
\acrodef{MOSFET}[MOSFET]{Metal Oxide Semiconductor Field-Effect Transistor}
\acrodef{MOS}[MOS]{Metal Oxide Semiconductor}
\acrodef{MRI}[MRI]{Magnetic Resonance Imaging}
\acrodef{NCS}[NCS]{Neuromorphic Cognitive Systems}
\acrodef{NDFSM}[NDFSM]{Non-deterministic Finite State Machine}
\acrodef{ND}[ND]{Noise-Driven}
\acrodef{NEF}[NEF]{Neural Engineering Framework}
\acrodef{NHML}[NHML]{Neuromorphic Hardware Mark-up Language}
\acrodef{NIL}[NIL]{Nano-Imprint Lithography}
\acrodef{NI}[NI]{Neural Interface}
\acrodef{NMDA}[NMDA]{\textit{N}-Methyl-\textsc{d}-aspartate}
\acrodef{NME}[NE]{Neuromorphic Engineering}
\acrodef{NN}[NN]{Neural Network}
\acrodef{NOC}[NoC]{Network-on-Chip}
\acrodef{NRZ}[NRZ]{Non-Return-to-Zero}
\acrodef{NSM}[NSM]{Neural State Machine}
\acrodef{OR}[OR]{Operating Room}
\acrodef{OTA}[OTA]{Operational Transconductance Amplifier}
\acrodef{PCB}[PCB]{Printed Circuit Board}
\acrodef{PCHB}[PCHB]{Pre-Charge Half-Buffer}
\acrodef{PCM}[PCM]{Phase Change Memory}
\acrodef{PC}[PC]{Personal Computer}
\acrodef{PDK}[PDK]{Process Design Kit}
\acrodef{PE}[PE]{Phase Encoding}
\acrodef{PFA}[PFA]{Probabilistic Finite Automaton}
\acrodef{PFC}[PFC]{Prefrontal Cortex}
\acrodef{PFM}[PFM]{Pulse Frequency Modulation}
\acrodef{PNI}[PNI]{Peripheral Nerve Interface}
\acrodef{PNS}[PNS]{Peripheral Nervous System}
\acrodef{PPG}[PPG]{Photoplethysmography}
\acrodef{PR}[PR]{Production Rule}
\acrodef{PSC}[PSC]{Post-Synaptic Current}
\acrodef{PSP}[PSP]{Post-Synaptic Potential}
\acrodef{PSTH}[PSTH]{Peri-Stimulus Time Histogram}
\acrodef{PV}[PV]{Parvalbumin}
\acrodef{PYR}[PYR]{Pyramidal}
\acrodef{QDI}[QDI]{Quasi Delay Insensitive}
\acrodef{RAM}[RAM]{Random Access Memory}
\acrodef{RA}[RA]{Resected Area}
\acrodef{RDF}[RDF]{Random Dopant Fluctuation}
\acrodef{RELU}[ReLu]{Rectified Linear Unit}
\acrodef{RISC}[RISC]{Reduced Instruction Set Computer}
\acrodef{RLS}[RLS]{Recursive Least-Squares}
\acrodef{RMSE}[RMSE]{Root Mean Square-Error}
\acrodef{RMS}[RMS]{Root Mean Square}
\acrodef{RNN}[RNN]{Recurrent Neural Network}
\acrodef{ROLLS}[ROLLS]{Reconfigurable On-Line Learning Spiking}
\acrodef{RRAM}[R-RAM]{Resistive Random Access Memory}
\acrodef{RSA}[RSA]{Respiratory Sinus Arrhythmia}
\acrodef{R}[R]{Ripple}
\acrodef{SAC}[SAC]{Selective Attention Chip}
\acrodef{SAT}[SAT]{Boolean Satisfiability Problem}
\acrodef{SCI}[SCI]{Spinal Cord Injury}
\acrodef{SCX}[SCX]{Silicon CorteX}
\acrodef{SD}[SD]{Signal-Driven}
\acrodef{SEM}[SEM]{Spike-based Expectation Maximization}
\acrodef{SHD}[SHD]{Spiking Heidelberg Digits}
\acrodef{SLAM}[SLAM]{Simultaneous Localization and Mapping}
\acrodef{SNN}[SNN]{Spiking Neural Network}
\acrodef{SNR}[SNR]{Signal to Noise Ratio}
\acrodef{SOC}[SoC]{System-On-Chip}
\acrodef{SOI}[SOI]{Silicon on Insulator}
\acrodef{SOZ}[SOZ]{Seizure Onset Zone}
\acrodef{SPI}[SPI]{Serial Peripheral Interface}
\acrodef{SP}[SP]{Separation Property}
\acrodef{SRAM}[SRAM]{Static Random Access Memory}
\acrodef{SST}[SST]{Somatostatin}
\acrodef{STDP}[STDP]{Spike-Timing Dependent Plasticity}
\acrodef{STD}[STD]{Short-Term Depression}
\acrodef{STP}[STP]{Short-Term Plasticity}
\acrodef{STT-MRAM}[STT-MRAM]{Spin-Transfer Torque Magnetic Random Access Memory}
\acrodef{STT}[STT]{Spin-Transfer Torque}
\acrodef{SVM}[SVM]{Support Vector Machine}
\acrodef{SW}[SW]{Software}
\acrodef{TCAM}[TCAM]{Ternary Content-Addressable Memory}
\acrodef{TFT}[TFT]{Thin Film Transistor}
\acrodef{TIME}[TIME]{Transverse Intrafascicular Multichannel Electrode}
\acrodef{TLE}[TLE]{Temporal Lobe Epilepsy}
\acrodef{UEA}[UEA]{Utah Electrode Array}
\acrodef{USB}[USB]{Universal Serial Bus}
\acrodef{USEA}[USEA]{Utah Slanted Electrode Array}
\acrodef{VHDL}[VHDL]{VHSIC Hardware Description Language}
\acrodef{VHSIC}[VHSIC]{Very High Speed Integrated Circuits}
\acrodef{VIP}[VIP]{Vasoactive Intestinal Peptide}
\acrodef{VLSI}[VLSI]{Very Large Scale Integration}
\acrodef{VNS}[VNS]{Vagal Nerve Stimulation}
\acrodef{VOR}[VOR]{Vestibulo-Ocular Reflex}
\acrodef{VSA}[VSA]{Vector Symbolic Architecture}
\acrodef{WCST}[WCST]{Wisconsin Card Sorting Test}
\acrodef{WTA}[WTA]{Winner-Take-All}
\acrodef{XML}[XML]{eXtensible Mark-up Language}
\acrodef{divmod3}[DIVMOD3]{Divisibility of a number by three}
\acrodef{hWTA}[hWTA]{Hard Winner-Take-All}
\acrodef{iEEG}[iEEG]{Intracranial Electroencephalography}
\acrodef{rSNN}[rSNN]{recurrent Spiking Neural Network}
\acrodef{sWTA}[sWTA]{soft Winner-Take-All}
\acrodef{smnist}[sMNIST]{sequential MNIST}
\acrodef{psmnist}[p-sMNIST]{permuted sequential MNIST}

\title{DendroNN: Dendrocentric Neural Networks for Energy-Efficient Classification of Event-Based Data}
\author[1,2]{Jann Krausse*}
\author[3]{Zhe Su*}
\author[4]{Kyrus Mama}
\author[3]{Maryada}
\author[5]{Klaus Knobloch}
\author[3]{Giacomo Indiveri}
\author[2]{Jürgen Becker}
\affil[1]{Infineon Technologies\\Dresden, Germany}
\affil[2]{Karlsruhe Institute of Technology\\Karlsruhe, Germany}
\affil[3]{Institute of Neuroinformatics\\University of Zurich and ETH Zurich, Zurich, Switzerland}
\affil[4]{Stanford University\\Stanford, CA, USA}
\affil[5]{Retired\\formerly Infineon Technologies\\Dresden, Germany}

\date{}

\begin{document}

\maketitle
\def\thefootnote{* }\footnotetext{These authors contributed equally to this work.}\def\thefootnote{\arabic{footnote}}

\abstract{
Spatiotemporal information is at the core of diverse sensory processing and computational tasks.
Traditional feed-forward spiking neural networks can be used to solve these tasks while offering potential benefits in terms of energy efficiency by using an event-based computing paradigm.
However, they have trouble decoding temporal information with high accuracy.
Therefore, they commonly resort to recurrence or delays to enhance their temporal computing ability which, however, bring major downsides in terms of hardware-efficiency.
In the brain, dendrites are computational powerhouses that just recently started to be acknowledged in such machine learning systems.
In this work, we specifically focus on a sequence detection mechanism present in branches of dendrites and translate it into a novel type of neural network by introducing a dendrocentric neural network, DendroNN.
DendroNNs identify unique incoming spike sequences as spatiotemporal features.
This work further introduces a rewiring phase to successfully train the non-differentiable spike sequences without the use of gradients.
During the rewiring, the network memorizes frequently occurring sequences and additionally discards those that do not contribute any discriminative information.
The resulting networks display competitive accuracies across various event-based time series datasets.
We also propose an asynchronous digital hardware architecture using a time-wheel mechanism that builds on the event-driven design of DendroNNs, eliminating per-step global updates typical of delay- or recurrence-based models.
By leveraging a DendroNN's dynamic and static sparsity along with intrinsic quantization, the architecture achieves up to 4× higher efficiency than state-of-the-art neuromorphic hardware at comparable accuracy on the same audio classification task, demonstrating its suitability for spatiotemporal event-based computing.
This work offers a novel approach to low-power spatiotemporal processing on event-driven hardware.
}

\section{Main}\label{sec:intro}



Temporal sequences are a fundamental currency of neural computation. 
Across sensory, cognitive, and motor systems, information is often encoded not merely in the presence of spikes, but in their precise spatiotemporal order and relative timing. 
Efficiently detecting such sequences remains a central challenge for artificial neural systems, particularly under constraints of locality, sparsity, and low power.

Classic deep neural networks treat time as a sequence of independent static frames, effectively ignoring temporal causality. 
Recurrent neural networks (RNNs) partially address this limitation by maintaining internal states that link events across time. 
However, they encode sequences implicitly through correlations between successive iterations rather than by directly computing temporal order. 
As a result, each new input requires global state updates, leading to rapidly increasing computational and energy costs with sequence length.

Spiking neural networks (SNNs) offer an event-driven alternative, but their basic computational units typically integrate inputs commutatively, summing spikes irrespective of arrival order.
Consequently, their responses reflect coincidence detection rather than true sequence sensitivity, limiting their ability to decode structured spatiotemporal patterns without external timing mechanisms or dense parameter updates.

In contrast, biological neurons represent events and behaviors as structured spike sequences distributed across space and time, enabling compact and efficient spatiotemporal representations through reproducible firing patterns~\cite{spike-based_strategies,polychronization}. 
In this context, dendrites have long been proposed to endow neurons with sequence selectivity, propagating depolarization toward the soma, and thereby triggering spiking, preferentially when synaptic inputs arrive in a consistent tip-to-soma order~\cite{rall1962electrophysiology}. Experimental work has directly observed such sequence dependence and has implicated depolarizing, voltage-dependent ion channels in its mechanism: inputs evoke larger inward currents through these channels when near an already depolarized dendritic region~\cite{branco2010dendritic}. More recent theoretical studies extend this view by highlighting hyperpolarizing, voltage-dependent channels as a complementary gating mechanism that suppresses indiscriminate passive spread of depolarization, allowing a synaptic input to advance a depolarized region only when it arrives nearby, so depolarization can \textit{step} toward the soma rather than diffuse broadly along the branch~\cite{boahen2022dendrocentric}. These results suggest that dendritic branches can operate as semi-independent sequence-detecting units that selectively respond to specific spatiotemporal patterns of synaptic activity.

Despite this insight, translating dendritic sequence computation into artificial systems presents several challenges. 
First, precisely simulating dendritic dynamics can quickly lead to an unsustainable hardware workload, spoiling all outlooks on an efficient system. 
Second, order-sensitive processing must be implemented in a fully event-driven manner, without requiring global neural state updates or synchronization.
Third, the combinatorial space of possible input sequences grows rapidly with input dimensionality and timescale, requiring mechanisms that can autonomously discover relevant temporal structure. 
Fourth, as the defining properties of spike sequences are inherently non-differentiable, optimization is nontrivial, requiring surrogate gradients or non-gradient-based learning rules~\cite{izhikevich2025spikingmanifesto}.

Current state-of-the-art systems have taken different approaches to decoding long-term temporal dependencies with \acp{SNN}.
Inspired by the heterogeneous signal transmission pathways in neural systems, delay-based \acp{SNN} keep track of temporal information by aligning timing of ingoing or outgoing signals through delay buffers.
The introduction of delays can achieve excellent accuracy across diverse datasets~\cite{conv_delays,snn_delaylearning,axon_delays}; however, it incurs significant hardware overhead, requiring either expensive shifting delay memory in digital implementations~\cite{loihi2} or large capacitors in mixed-signal implementations~\cite{denram}.
Additionally, recurrence has been a proven method to improve the temporal computing capabilities of neural networks.
This is also true for \acp{SNN}, as added recurrence improves the accuracy across multiple time-series benchmarks~\cite{eprop}.
However, the resulting accuracies are often inferior to those achieved by networks employing delays, while the additional recurrence matrix imposes a substantial burden in both energy consumption and area cost~\cite{reckon,elfcore}.

Here, we introduce a dendritic sequence detection model called DendroNN that abstracts dendritic branches as compartmentalized processing units capable of detecting structured spike sequences.
Presynaptic neurons encoding distinct elements of a sequence form connections with specific dendritic compartments, effectively mapping temporal structure onto spatial organization. 
A neural unit emits an output event only when incoming spikes arrive in the correct order and within defined temporal windows, while deviations in timing or ordering suppress activation.
Moreover, the ability to process multiple sequences in parallel across distinct dendritic branches allows a single unit to multiplex temporal computations without interference.
This mechanism enables robust detection of temporally precise patterns while naturally rejecting irrelevant, noisy, or incomplete inputs.


We shift computation away from global synaptic weight tuning toward reordering sparse binary connections and carefully adapting temporal intervals while leveraging high degrees of dynamic and static sparsity; a formulation based on operations that require minimal load.

The model's definition imposes strong binarization and sparsity constraints that make it difficult to train its hidden layers with gradient-based approaches.
To resolve this conundrum, we deploy a gradient-free rewiring phase that identifies spike sequences among the inferred data that are rich in information.
Our models display competitive performance across multiple datasets when compared to delay-based and recurrent \acp{SNN} while consuming comparably little memory footprint.

We further propose a near-memory asynchronous digital computing architecture that operates without a clock \cite{selftime}. The design is built around a time-wheel mechanism that exploits the event-driven nature of DendroNN while eliminating the per-time-step global updates typical of delay- or recurrence-based designs.
As a result, memory access costs scale with event sparsity rather than the length of the time window. 
The hardware is implemented in the GlobalFoundries\textsuperscript{\textregistered}~22FDX FDSOI technology node and evaluated using post-layout simulations. 
By leveraging DendroNN’s high dynamic and static sparsity as well as its intrinsic quantization, the architecture achieves up to a $4\times$ efficiency improvement over state-of-the-art neuromorphic hardware at comparable accuracy on the same audio classification task, demonstrating strong efficiency for spatiotemporal event-based computation.

With this, we present an end-to-end training and deployment scheme for our novel spike-based neural network.
This underlines the continuing potential of sparse systems when accelerated on dedicated event-based and asynchronous hardware, fulfilling the promise of neuromorphic computing.

\section{Results}\label{sec:results}
\subsection{DendroNN Model}
\label{sec:model}
\subsubsection{Sequence Detection Abstraction}
Instead of modeling complex ionic currents, the DendroNN unit abstracts dendritic sequence detection along a single branch by emitting a single binary output spike when a well-defined spike sequence appears in the input data. 
A spike sequence as a spatiotemporal feature is defined by the following properties
\begin{enumerate}
    \item number of spikes $N_\text{S}$ in sequence,
    \item presynaptic neuron indices defining the spatial origins of spikes in the sequence $X=[ x_0, x_1, \ldots, x_{N_\text{S}-1} ]$,
    \item temporal order of spikes, which is a permutation $\sigma_\text{S}\in S_{N_\text{S}}$,
    \item inter-spike intervals between consecutive spikes $\{\Delta t_1, \Delta t_2, \ldots, \Delta t_{N_\text{S}-1}\}$.
\end{enumerate}
Every unit in the DendroNN is sensitive towards a single such sequence, with one dedicated spine per spike within the sequence.
Similar to its function in the biological sequence detection, a spine presents one segment of a DendroNN unit dedicated to detecting a single spike within the complete sequence.
For each unit, the above parameters define the respective sequence.
As each unit will only be activated by spikes that are part of that specific sequence, all of its input connections are given by $X$, the spatial origin of spikes within the sequence.
Moreover, since the detection of a sequence and all its contributing spikes demands no further transformation of the input, all connections forwarding spikes to a unit can be binary.

During inference, the units expect spikes in the correct sequential order $\sigma_\text{S}$.
This is realized via the spatial organization of spines within the unit: the spatially first spine expects the temporally first spike.
Furthermore, after the reception of single spikes within the sequence, the unit expects the next spike after the respective inter-spike interval $\Delta t_i$.
As there is neither a higher nor a lower limit for possible $\Delta t_i$, units can capture features across rapid, as well as drawn out timescales.
In fact, even a single feature, i.e. a singular sequence of spikes, can compile temporal dependencies across multiple timescales if the $\Delta t_i$ are different by large factors.
This promises a high capability in decoding temporal information.
Depending on the data, it might make sense to allow each spike to arrive within a certain time window $\Delta T$ around $\Delta t_i$ rather than at exactly $\Delta t_i$.
This spike acceptance window is a global hyperparameter, contrasting the above sequence properties unique to each unit.
For implementation details, please refer to Section \ref{sec:methods:model}.

For a specific unit, we define the ordered list of spike-origins as $\tilde{X}$ by applying the unit's temporal order-defining permutation $\sigma_\text{S}$ to its $X$: 
\begin{equation}
    \tilde{X} = \sigma_\text{S}(X) =
    [ x_{\sigma_\text{S}(0)}, x_{\sigma_\text{S}(1)}, \ldots, x_{{\sigma_\text{S}(N_\text{S}-1)}} ] = [ \tilde{x}_0, \tilde{x}_1, \ldots, \tilde{x}_{N_\text{S}-1} ].
\end{equation}
In essence, spike origins in $\tilde{X}$ are now listed in temporal order, according to the sequence of spikes.
Thus we can define the DendroNN activation function given a time-dependent input signal $I_t(x)$ as
\begin{equation}
    o_{t_0}=\bigwedge_{i=0}^{N_S-1}I_{t_i}(\tilde{x}_i),\quad t_i = t_0-\sum_{j=0}^{i}\Delta t_j,
\label{eq:activation_function}
\end{equation}
with $\Delta t_{0}=0$.
Equation \eqref{eq:activation_function} essentially ANDs all inputs $I$ from the correct spatial origins $\tilde{x}_i$ and at the correct times $t_i$.
Hence, if the correct sequence occurs the unit will emit an output of value one, else the output is zero.

\begin{figure}[ht]
    \centering
    \includegraphics[width=0.8\textwidth]{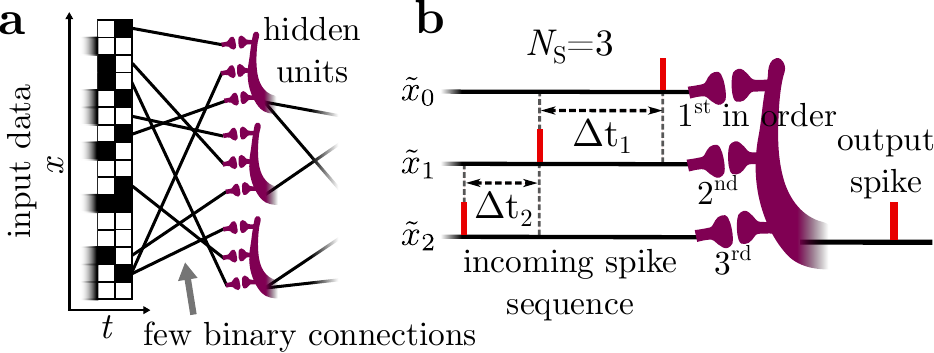}
    \caption{
    \textbf{(a)} Schematic of the network architecture. Sparse, binary connections route input spikes $x$ to a set of hidden units, each tuned to a specific spatiotemporal pattern. \textbf{(b)} Example of sequence detection for $N_s = 3$. A hidden unit responds only when spikes arrive from the correct spatial origins $\tilde{x_0},\tilde{x_1},\tilde{x_2}$ in the prescribed temporal order, with inter-spike intervals $\Delta t_1$ and $\Delta t_2$. Inputs that are out of order or mistimed are suppressed. When the full target sequence is observed, the unit emits a single output spike.  }
    \label{fig:unit_properties}
\end{figure}

\begin{figure}[ht]
    \centering
    \includegraphics[width=0.8\textwidth]{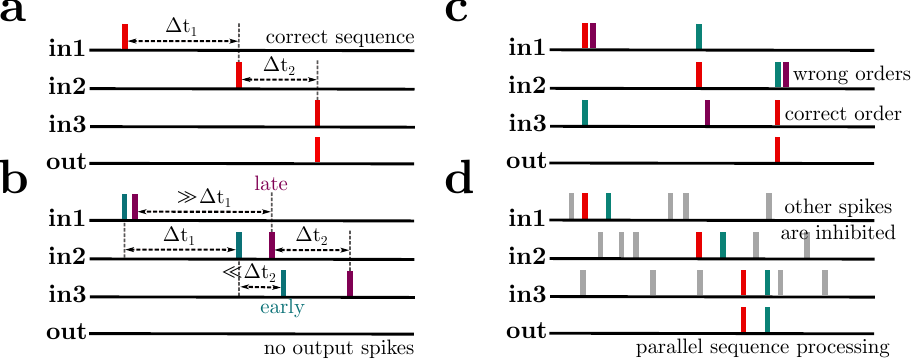}
    \caption{Exemplary input spike patterns and model responses. \textbf{(a)} All input spikes arrive from the correct spatial channels and in the prescribed temporal order, with the required inter-spike intervals $\Delta t_1$ and $\Delta t_2$, resulting in the emission of an output spike. \textbf{(b)} Although the spike order is preserved, deviations in the inter-spike intervals (early or late arrivals) violate the temporal constraints and suppress the output response. \textbf{(c)} Only the spikes belonging to the correctly ordered sequence (red) satisfy the spatiotemporal constraints; spikes from incorrectly ordered sequences do not elicit an output. \textbf{(d)} Multiple sequences can be processed in parallel: irrelevant spikes are inhibited, while all valid target sequences are reliably detected and trigger output events.}
    \label{fig:example_seqs}
\end{figure}

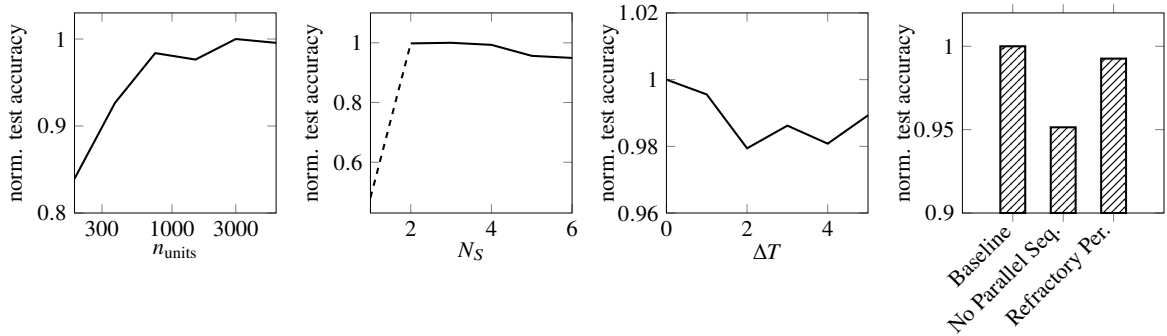
\begin{figure}
    \centering
    \begin{tikzpicture}
\begin{groupplot}[
    group style={
        group size=4 by 1,
        horizontal sep=1.6cm,
        vertical sep=1.1cm,
    },
    width=5cm,
    height=5cm,
    enlargelimits=false,
    every axis y label/.style={at={(axis description cs:-0.28,0.5)}, rotate=90, anchor=center},
    grid=none,
    every axis x label/.style={at={(axis description cs:0.5,-0.19)}, rotate=0, anchor=center},
    every axis plot/.append style={line width=1pt},
    no markers,
    ylabel={norm. test accuracy},
]

\nextgroupplot[xlabel={$n_\text{units}$}, xmode=log, ymin=0.80, ymax=1.03, xtick={300, 1000, 3000}, xticklabels={300, 1000, 3000}]
\addplot+[solid, black] coordinates {(187,0.8394) (375,0.9265) (750,0.9838) (1500,0.9764) (3000,1.0) (6000,0.9956)};

\nextgroupplot[xlabel={$N_S$}, ymin=0.43, ymax=1.1]
\addplot+[solid, black] coordinates {(2,0.998) (3,1.0) (4,0.9931) (5,0.9563) (6,0.9493)};
\addplot+[dashed, black] coordinates {(1,0.4813) (2,0.998)};

\nextgroupplot[xlabel={$\Delta T$}, ymin=0.96, ymax=1.02]
\addplot+[solid, black] coordinates {(0,1.0) (1,0.99556) (2,0.9794) (3,0.9862) (4,0.9808) (5, 0.9893)};

\nextgroupplot[
    ybar,
    bar width=12pt, 
    xtick={1, 2, 3}, 
    xticklabels={Baseline, No Parallel Seq., Refractory Per.},
    xticklabel style={rotate=45, anchor=east},
    ymin=0.9, ymax=1.02, 
    xmin=0, xmax=4,
]
\addplot[black, pattern=north east lines] coordinates {(1, 1.0) (2, 0.9514) (3, 0.9926)};

\end{groupplot}
\end{tikzpicture}
    \caption{Visualization of the impact of different model parameterizations on its classification performance.}
    \label{fig:model_hparams}
\end{figure}

Figure \ref{fig:unit_properties} visualizes the working principle of the DendroNN unit and summarizes its central properties. 
Additionally, Figure \ref{fig:example_seqs} exemplifies possible input spikes and the corresponding response of a DendroNN unit to that input. 
Note that, as illustrated in Figure \ref{fig:example_seqs}\textbf{d}, a unit can detect multiple instances of the same sequence in parallel.
This allows units to ignore unimportant or noisy spikes while filtering out the sought after sequences. 
Empirical results further underline the importance of this property (see Figure \ref{fig:model_hparams}). 

Adding to the above properties, it is possible to impose a refractory period onto a unit after they have spiked.
Here, we only consider refractory periods longer than the input data sequence, limiting a unit's output to a maximum of one spike per inference.
While this does not add value in terms of classification performance, it significantly decreases computational demand, as each emitted spike adds to the total energy and latency cost of the target hardware.
The influence of refractory periods on the models classification is shown in Figure~\ref{fig:model_hparams}.

Stacking multiple hidden layer is possible and easily implemented.
By doing so, succeeding layers combine subsequences detected by preceding layers into longer sequences, introducing a form hierarchical organization of these subsequences.
Considering $l$ layers with $n_{\text{units},i}$, $n_\text{in}$ input channels, and a total sequence length of $N_{S,\text{tot}}=\sum_{i=1}^lN_{S,i}$, this reduces the search space from $n_\text{in}^{\sum_{i=1}^lN_{S,i}}$ to $\prod_{i=0}^{l-1}n_{\text{units},i}^{N_{S,i+1}}$ with $n_{\text{units},0}=n_\text{in}$.
In this work, because considered sequence lengths are relatively short, we only utilize networks that comprise a single hidden layer.

Finally, the hidden layer is connected to a fully connected linear layer, delivering weighted binary spikes resulting from detected sequences to a layer of $n_\text{classes}$ integrator units. 
Classification accuracy and cross-entropy loss are computed at this final integrator layer, from which gradients are backpropagated.

\subsubsection{Compatibility with Event-Driven Digital Hardware}
Whether realized on digital hardware operating in discrete time~\cite{spinnaker, truenorth, loihi} or on mixed-signal platforms operating in continuous time~\cite{neurogrid, braindrop, dynapse}, neuromorphic systems are typically designed in an asynchronous, i.e., clockless or self-timed, fashion. This design paradigm naturally aligns with the event-driven behavior of SNN. 

Despite these benefits, existing asynchronous digital neuromorphic hardware continues to rely heavily on vector-based operations that are not inherently event-driven and whose execution time scales with the length of the simulation window. Such low-arithmetic-intensity operations, including neuron membrane potential updates and eligibility trace computations, incur substantial memory access overhead, thereby reducing the efficiency advantages typically associated with SNNs.

In contrast to conventional approaches, the unit activation function of DendroNN enables a fully end-to-end event-driven hardware implementation, as shown in Equation~\eqref{eq:activation_function}. Only two low-cost element-wise operations are required. The first tracks running inter-spike intervals to enforce the correct temporal spacing between consecutive spikes. The second detects pairwise spike occurrences once the corresponding inter-spike interval has elapsed. While the latter operation is naturally event-driven, the former is implemented using a time-wheel mechanism (described in Section~\ref{sec:hw}), which significantly reduces memory access overhead.

Finally, since units emit non-zero outputs only upon successful sequence detection, the resulting temporal and spatial sparsity is inherently well suited to event-driven hardware architectures.


\subsubsection{Model Compression}
Concerning the DendroNN's hidden layer, assuming $n_\text{in}$ input channels and $n_\text{hidden}$ hidden units, having only $N_\text{S}$ input connections per unit, i.e., one connection per spine, yields a inherent spatial sparsity of $S_\text{spat}=\frac{N_\text{S}\cdot n_\text{hidden}}{n_\text{in}\cdot N_\text{S}\cdot n_\text{hidden}}=\frac{1}{n_\text{in}}$ and a total number of $N_\text{S}\cdot n_\text{hidden}$ hidden connections.
Furthermore, each of those connections carries a binary value.
As seen by the choice of network topologies below (see Section~\ref{sec:methods:hparams}), the number of hidden units tends to be 5-10$\times$ larger than for other types of neural networks.
For most datasets, the high degree of spatial sparsity in addition to deploying a single hidden layer still leads to a favorable memory footprint of the hidden connections.
However, in many cases, the large number of hidden units leads to the linear output layer being a dominating contribution to the total memory footprint.
To reduce the rather large memory footprint of the output layer, we utilize magnitude pruning and quantization.
Fortunately, since we only target the linear output projection, pruning levels can reach high values in combination with quantization to int8 values without heavily impacting classification accuracy.

\subsection{Rewiring Phase}\label{sec:rewiring_phase}
While the DendroNN unit exhibits extraordinary potential for acceleration on digital hardware, most of its central properties are non-differentiable.
Crucially, this impacts the plasticity of parameters defining a unit's sequence, e.g., integer-valued inter-spike intervals $\Delta t_i$ and the spike origins $x_i$ which are realized via a set of $N_S$ binary connections.
However, the number of combinatorially possible sequences $N_\text{seqs}$ quickly explodes.
A hypothetical dataset with a number of input channels and an input sequence length of 100 each would then quickly yield $2.5\cdot10^{9}$ individual sequences assuming the networks computes on sequences of 3 spikes (see Equation~\ref{eq:num_possible_seqs}).
As surely there exists a lot of redundancy in all $2.5\cdot10^{9}$ sequences, we assume the maximum number of hidden units in order to satisfy hardware resource constraints to be many orders of magnitudes lower.
This demonstrates the need for the ability to adapt the DendroNN's sequences based on the given data.
To solve this infeasability, we devised a rewiring phase that adapts a network's units based on observed spatiotemporal patterns in the input data without the calculation of gradients.

\subsubsection{Finding Common Sequences}
Addressing the issue of how to find important sequences without being able to calculate gradients on the sequence-defining properties, the central realization is that potentially useful features are sequences that are commonly observed in the given data.
Based on that idea, we deploy a rewiring phase inspired by the sparse binary coincidence memories developed in \cite{hopkins2023bitbrain, hopkins2018spiking}.
During that pre-supervised training phase, we infer the network with training samples until a target number of important sequences is found.
For each unit, we define a so-called \emph{longevity state} which altered depending on the occurrence of the sequence that unit is currently trying to detect.
Then, for every sample the network is inferred with, the longevity of each unit is altered by a reward or penalty amount depending on whether the respective sequence appeared within the data sample or not.
 
Throughout the inference of many samples, the longevity of all units will either drift towards the positive regime or the negative regime.
If a unit's longevity is sufficiently high, past a positive threshold, it is frozen and consolidated. 
Alternatively, if a unit's longevity is too low, past a negative threshold, the unit is rewired: all non-global sequence-defining parameters are randomly redrawn and its longevity is reset to zero.

\begin{figure}[ht]
    \centering
    \includegraphics[width=0.8\textwidth]{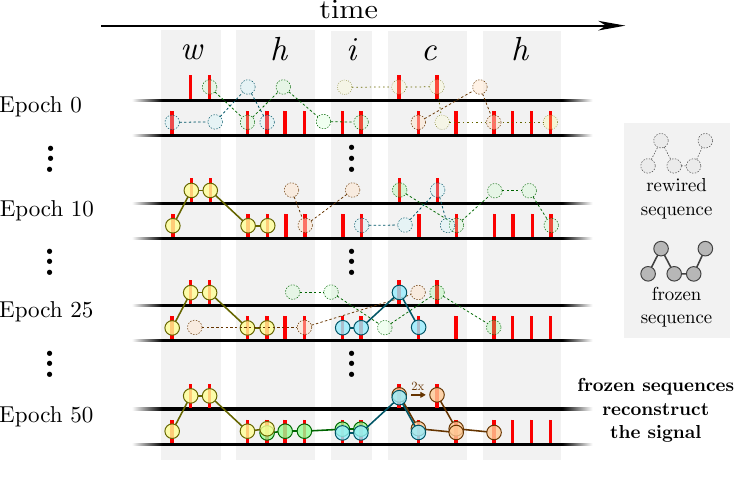}
    \caption{Visualization of the rewiring phase using the NeuroMorse train set sample "which" as an example. Throughout multiple epochs, initially randomly arranged sequences are rewired and frozen to match existing spatiotemporal patterns in the sample.}
    \label{fig:rewiring-phase}
\end{figure}

\begin{figure}[ht]
    \centering
    \includegraphics[width=0.5\textwidth]{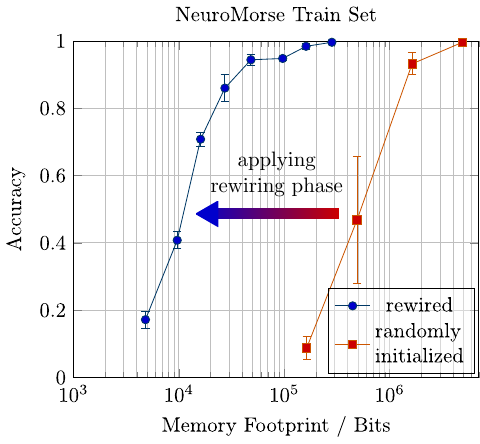}
    \caption{Impact of applying the rewiring phase regarding the model's footprint. In the case of the NeuroMorse dataset, network size can be reduced by 10-100x without suffering accuracy loss. This effect is drastically amplified for datasets of higher spatial and temporal complexity.}
    \label{fig:rewiring-phase-advantage}
\end{figure}

Figure~\ref{fig:rewiring-phase} visualizes the process of finding spike sequences that match common spatiotemporal patterns in the input data. 
The chose sample is the word "which" converted to morse code as part of the NeuroMorse dataset. 
While the four monitored sequences of different length initially seem to be spatiotemporally randomly arranged, throughout multiple epochs they are rewired and frozen one after another to match re-occurring patterns found in the sample data.
The actual advantage of the rewiring phase is demonstrated in Figure~\ref{fig:rewiring-phase-advantage}. 
It compares the NeuroMorse train set accuracy of DendroNNs of different sizes with and without utilizing rewiring.
As seen there, rewiring the units reduces the network size by one to two orders of magnitude.
Note that, while reducing the network size by 10-100x is a great benefit, NeuroMorse train data is highly simplistic, consisting of only two input channels and a consistent time difference between individual spikes.
For such data, according to Equation~\eqref{eq:num_possible_seqs}, the combinatorial number of possible sequences is rather low which decreases the potential gain through unit rewiring.
Other datasets, however, quickly reach multiple hundred input channels as well as a much higher temporal variance in event timings, leading to a drastically larger potential benefit by deploying the rewiring phase for more complex data.

\subsubsection{Handling Non-Selective Sequences}
As written above, this rewiring continues until a target number of good sequences are found.
Evaluating the relative occurrence of those sequences with respect to all data classes, however, reveals that sequences found via this approach commonly appear in all samples of all classes.
Such sequences, when used for differentiating samples of different classes, do not reduce the entropy of the target classes and thus have zero informational gain making them as useful for classification as a feature that appears in no sample.
Therefore, there is a need to check the selectivity of potentially frozen units towards subsets of classes before permanently choosing them for the supervised training that follows the rewiring phase.

In order to verify that each frozen unit is indeed selective towards a subset of classes, we deploy a quick validation in which we count total number of emitted spikes of each unit for every class.
From there we deduce probabilities that, if a given unit spikes, the data sample is of a certain class.
Finally, all units whose probabilities meet a pre-defined selectivity criterion are permanently frozen.
For implementation details, see Section~\ref{sec:methods:rewiring_phase}.
Such criteria can practically impose any probability distribution on the occurrence of a sequence across available classes.
Crucially, though, it enforces that, for each class in the dataset, there exist sequences that represent those classes with a given probability.
\subsection{Algorithmic Experiments}
\label{sec:experiments}
\subsubsection{NeuroMorse}
\label{sec:experiments:neuromorse}
The NeuroMorse dataset consists of two channel event-based data, converting the Morse code of English words into spike trains \cite{neuromorse}.
For that, the dots and dashes of the respective Morse code represent a spike each and are spaced in time using pre-defined inter-spike intervals.
The train set comprises the 50 most frequent words in the English language.
Interestingly, the test set is a compilation of words included in 50\,441 Wikipedia articles.
This means that the large majority, roughly two thirds, of the test set consists of null class samples.

It is well visible that standard \acp{SNN} do very poor on the noise-free data, further worsening after noise is applied.
Two things are important here.
Firstly, the baseline accuracy of networks trained on noise-free data is low primarily because the \acp{SNN} deployed by the authors cannot handle null class samples.
Secondly, the \acp{SNN} suffer quickly from applied noise.
It should be noted that even in the absence of those two problems the performance would not be perfect, as the reported train set accuracy of \acp{SNN} is roughly 80\%, highlighting the ``limited ability to learn temporal hierarchies in current spiking architectures'' while having temporal causality~\cite{neuromorse}.

Although there exist no works benchmarking this dataset on other network architectures as it was published relatively recently, we find that the dataset presents a compelling example to showcase some interesting properties of DendroNNs.
Table~\ref{tab:results_neuromorse} compares DendroNNs with SNNs on different noise types and intensities. The choice of hyperparameters is compiled and discussed in Section~\ref{sec:methods:hparams}.

Contrasting the poor results of \acp{SNN}, the DendroNNs achieve rather high accuracies, with 90.22\% accuracy on the noise free data and >45\% for all levels of added noise.
This is a first sign of rich sequence decoding capabilities of the DendroNN but more than anything it demonstrates an interesting uncertainty detection feature of these networks.
If DendroNNs do not detect any sequences learned during training, they emit rather few events, which can be used to detect null class samples.
For details, see Section~\ref{sec:methods:nullclass}.

\begin{table}[h]
    \centering
    \caption{Benchmarking of DendroNN and \ac{SNN} on the NeuroMorse dataset for classification of spike sequences.}

\resizebox{\textwidth}{!}{ 
\begin{tabular}{l c ccc ccc ccc}
\toprule
\multirow{2}{*}{\textbf{Poissonian}} & \textbf{Jitter} & \multicolumn{3}{c}{None} & \multicolumn{3}{c}{Low} & \multicolumn{3}{c}{High} \\
\cmidrule(lr){3-5} \cmidrule(lr){6-8} \cmidrule(lr){9-11}
& \textbf{Dropout} & None & Low & High & None & Low & High & None & Low & High \\[0.1cm]
& \textbf{Model} & & & & & & & & & \\
\midrule
\multirow{2}{*}{None} & SNN \cite{neuromorse} & 12.27\% & 4.37\% & 2.48\% & 6.22\% & 2.49\% & 1.69\% & 5.70\% & 2.07\% & 1.66\% \\
& \textbf{DendroNN*} & 90.22\% & 81.53\% & 75.18\% & 51.63\% & 51.55\% & 51.82\% & 52.29\% & 50.24\% & 51.94\% \\[0.1cm]
\multirow{2}{*}{Low} & SNN \cite{neuromorse} & 11.81\% & 4.30\% & 2.40\% & 6.26\% & 2.42\% & 1.65\% & 5.76\% & 2.00\% & 1.60\% \\
& \textbf{DendroNN*} & 73.24\% & 68.79\% & 66.24\% & 51.79\% & 52.34\% & 50.55\% & 49.58\% & 50.85\% & 49.56\% \\[0.1cm]
\multirow{2}{*}{High} & SNN \cite{neuromorse} & 11.34\% & 4.09\% & 2.30\% & 6.16\% & 2.35\% & 1.70\% & 5.84\% & 2.06\% & 1.57\% \\
& \textbf{DendroNN*} & 67.01\% & 64.65\% & 61.60\% & 50.26\% & 49.28\% & 49.67\% & 48.99\% & 46.21\% & 47.06\% \\
\bottomrule
\multicolumn{6}{l}{* This work.}

\end{tabular}
}

    \label{tab:results_neuromorse}
\end{table}

\subsubsection{Sequential and Permuted Sequential MNIST}
\label{sec:experiments:mnist}
The \ac{smnist} dataset presents a standard benchmark for time series classification.
It is creating by flattening the original 28x28 handwritten digit image and treating the resulting 784x1 tensor as a time series.
While \ac{smnist} is a rather easily solvable problem for many networks, it's permuted counterpart poses a rather tough challenge \cite{adaptive_srnn}.
To yield \ac{psmnist}, the same permutation is applied to the time series in all samples of \ac{smnist}.
Since the permutation distorts information that was originally within a close temporal proximity, the difficulty of solving this dataset increases for most models.
Importantly, pixels in the unaltered MNIST samples carry greyscale values between 0 and 255.
Since the DendroNN model processes binary events only, we quantize the pixel values so that any greyscale value greater than 0 is represented by the value 1.

Table~\ref{tab:results_mnist} summarizes the performance of the DendroNNs and other spiking models on both datasets.
Again, the results showcase the poor sequence decoding ability of basic \acp{SNN}, as they do not break out of the random chance regime.
While the DendroNNs struggle to break into state of the art regime, they easily compete with other spiking models while showcasing a potentially much lower memory footprint.
This again underlines the decent time series computing capability of DendroNNs.
The choice of hyperparameters can be found in Section~\ref{sec:methods:hparams}.
\phantom{\cite{lsnn_bellec,delrec}}  

\begin{table}[ht]
    \centering
    \caption{Benchmarking of DendroNN and other spiking models on \ac{smnist} and \ac{psmnist} for classification of time sequences. DendroNN results are averaged over 5 runs. Memory footprints are calculated using Equation~\eqref{eq:memory}. Memory sizes of references assume 32\,bit parameters and do not include contribution of hidden and output unit states as well as delay buffers to the total memory footprint, the latter of which is expected to contribute significantly.}

\begin{tabular}{llcc}
\toprule
\textbf{Dataset} & \textbf{Model} & \textbf{Accuracy} & \makecell{\textbf{Memory}\\\textbf{Footprint / MBit}}\\

\midrule

\multirow{5}{*}{sMNIST} & SRNN \cite{adaptive_srnn} & 10.00\%$^+$ & 1.42 \\
& Adaptive SRNN \cite{adaptive_srnn} & 97.82\% & 1.42 \\ 
& LSNN (L2L+DeepR) \cite{lsnn_bellec} & 96.40\% & 0.72 \\  
& \textbf{DendroNN*} & $97.50\pm0.11$\% & 1.71 \\
& \textbf{DendroNN* (compressed)} & $96.23\pm0.15$\% & 1.27 \\

\midrule

\multirow{5}{*}{p-sMNIST} & SRNN \cite{adaptive_srnn} & 10.00\%$^+$ & 2.15 \\
& Adaptive SRNN \cite{adaptive_srnn} & 91.00\% & 1.42 \\
& DelRec (delay-based SNN) \cite{delrec} & 96.21\% & 5.12$^\dagger$ \\  
& \textbf{DendroNN*} & $94.88\pm0.08$ & 1.71 \\  
& \textbf{DendroNN* (compressed)} & $93.95\pm0.05$ & 1.27 \\  

\bottomrule
\multicolumn{2}{l}{* This work, $^\dagger$ excluding size of delay buffers, $^+$ random chance.}

\end{tabular}

    \label{tab:results_mnist}
\end{table}

\subsubsection{Spiking Heidelberg Digits}
\label{sec:experiments:shd}
\ac{SHD} is an event-based audio classification dataset and a popular choice for benchmarking of spiking models \cite{shd}.
Recently, the possibility of learning synaptic delays in \acp{SNN} has led to strong improvements in model performance on this task, leading to delay-based networks presenting the current state of the art.

The density of spikes in \ac{SHD} recordings is so dense that the attempt to pre-process them into time series with reasonable bin size, i.e., $\ge$1\,ms, leads to the binning of multiple spikes into the same bin.
This results in samples containing events of values larger than one.
Moreover, we believe that these events and their value are of specific importance as they indicate the underlying high density of incoming spikes.
However, as pointed out above, the DendroNN takes only binary event streams as input.
This leads to the major issue of enabling DendroNNs to process events with values larger than one.
To solve this, the data is pre-processed so that the event values are contained in the binarized data, e.g., by translating value information into spatial information.
For implementation details see Section~\ref{sec:methods:data}.
We want to point out that the representation of events of larger values as spikes possibly carries the risk of losing generalizability of the model along the event value axis.

Table~\ref{tab:results_SHD} compares the DendroNN with other spiking models on the \ac{SHD} dataset.
While the DendroNN shows decent classification performance and easily outperforms basic recurrent \ac{SNN} topologies, its performance is inferior to those of delay-based \acp{SNN}.
We attribute this gap to the initially high quantization and sparsity of the model, the limited ability of the rewiring phase to identify the right sequences as important spatiotemporal features, and the potential loss in generalizability due to the conversion of events with value larger than one into spikes.
Notably, DendroNN networks display a much smaller memory footprint than the other models.
Contrasting recent trends of validating on \ac{SHD}'s test set, our results are trained with validation on dedicated 20\% of the train set, similar to all listed references.
All hyperparameters can be found in Section~\ref{sec:methods:hparams}.
\phantom{\cite{adlif}}  

\begin{table}[ht]
    \centering
    \caption{Benchmarking of DendroNN other spiking models on the \ac{SHD} dataset for classification of event-based data. DendroNN results are averaged over 5 runs. Memory footprints are calculated using Equation~\eqref{eq:memory}. Memory sizes of references assume 32\,bit parameters and do not include contribution of hidden and output unit states as well as delay buffers to the total memory footprint, the latter of which is expected to contribute significantly.}
    
\begin{tabular}{lcc}
\toprule
\textbf{Model} & \textbf{Accuracy} & \makecell{\textbf{Memory}\\\textbf{Footprint / MBit}}\\

\midrule

RSNN \cite{snn_delaylearning} & 85.60\% & 8.00 \\  
SNN w/ random dend. delays \cite{denram} & 90.1\% & 7.17$^\dagger$ \\
RSNN w/ learnable syn. delays \cite{snn_delaylearning} & 93.2\% & 40.04$^\dagger$ \\ 
adLIF \cite{adlif} & 93.79\% & 14.40 \\  
\textbf{DendroNN*} & $89.24\pm0.31$\% & 0.60 \\
\textbf{DendroNN* (compressed)} & $88.48\pm0.25$\% & 0.50 \\

\bottomrule
\multicolumn{2}{l}{* This work, $^\dagger$ excluding size of delay buffers.}

\end{tabular}

    \label{tab:results_SHD}
\end{table}
\subsection{Asynchronous Hardware Implementation}
\label{sec:hw}

\begin{figure}[ht]
    \centering
    \includegraphics[width=0.95\textwidth]{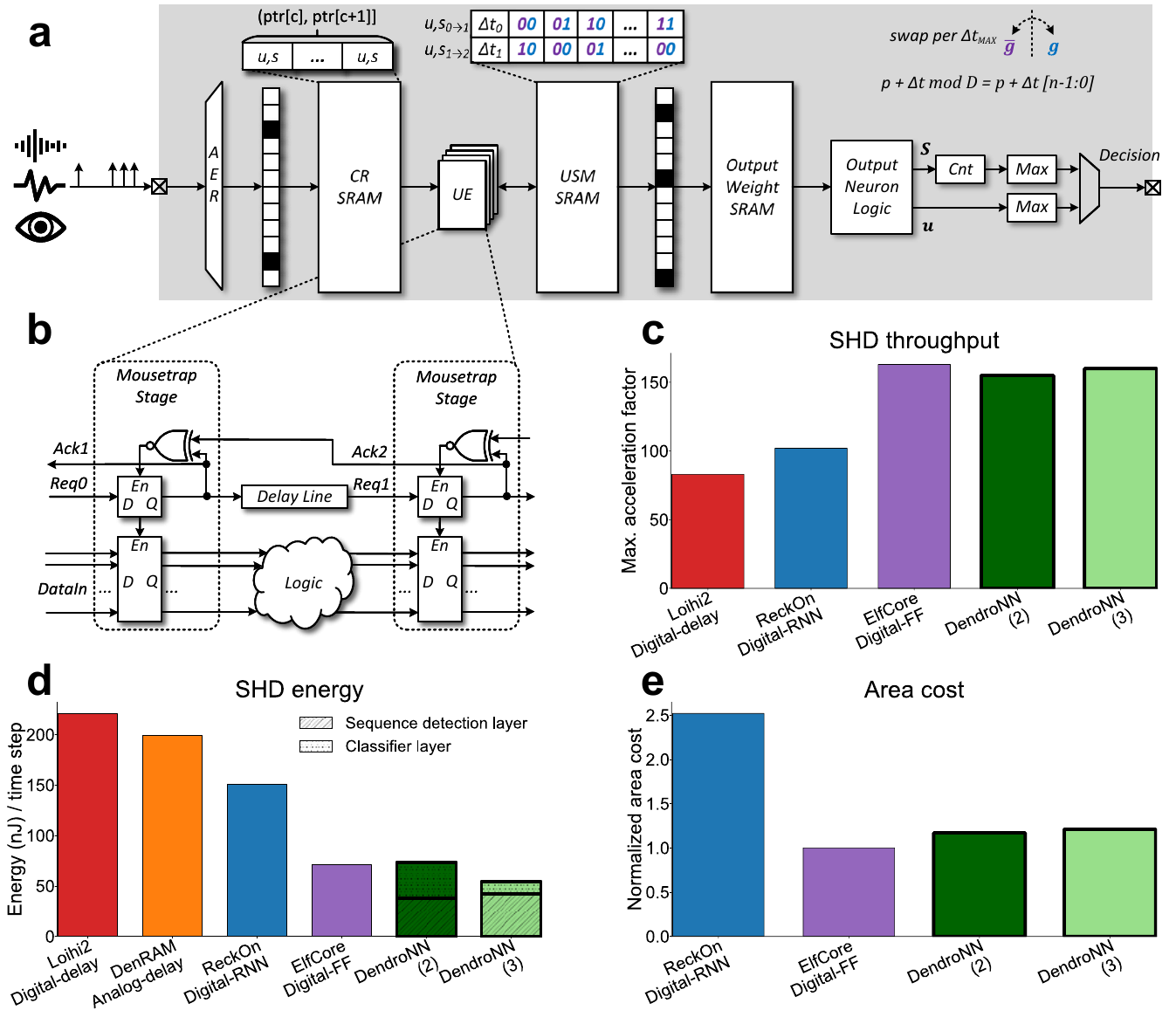}
\caption{A near-memory asynchronous digital computing architecture for DendroNN, along with a comparison of hardware metrics against other aggressively optimized platforms achieving comparable accuracy on the same task.
(a) Hardware microarchitecture. Event data from the sensor are decoded into channel addresses using Address Event Representation (AER). Each entry in the connectivity router (CR) memory stores four sparse orthogonal indices corresponding to the postsynaptic unit and spine. Four parallel unit update engines (UEs) handle due-condition detection, new spike scheduling, and weight update operations. Spikes generated by the hidden units are propagated to the output classifier layer, which supports both spike-count--based decisions and neuron membrane potential--based decision making.
(b) The \textit{Mousetrap} asynchronous pipeline~\cite{mousetrap} is employed throughout the AER, UE, and output neuron logic. This pipeline offers minimal hardware cost and high operating speed. By eliminating the need for a global clock, it naturally aligns with the event-driven characteristics of DendroNN.
(c) DendroNN (3) achieves up to $1.9\times$ higher throughput than delay-based hardware, while providing comparable throughput to feedforward networks with significantly higher accuracy.
(d) DendroNN (3) delivers up to $4.1\times$ higher energy efficiency than delay-based hardware and $2.8\times$ higher energy efficiency than recurrence-based hardware, primarily due to its fully event-driven architecture.
(e) DendroNN achieves $2\times$ higher area efficiency than recurrence-based hardware, mainly due to its binary weights and efficient spike-activity storage.}
\label{fig:dendronn_hardware}
\end{figure}

\subsubsection{Microarchitecture Overview}
\label{sec:hw_overview}
We target an inference-time hardware realization of DendroNN units specialized to $N_S = 3$ spines with an exact spike-acceptance window $\Delta T = 0$. This specialization results in two ($N_S - 1$) state stages per unit. As illustrated in Fig.~\ref{fig:dendronn_hardware}(a), the accelerator is organized as an event-driven pipeline comprising: (i) an Address Event Representation (AER) interface that bins the incoming event stream into tuples $(t, c)$ denoting time-bin and channel indices; (ii) a Connectivity Router (CR) that maps each spiking channel to its corresponding unit--spine targets; (iii) an on-chip Unit-State Memory (USM) that maintains per-unit temporal state; (iv) multiple parallel Update Engines (UEs) that perform due checks, state updates, and spike emission (four shown in Fig.~\ref{fig:dendronn_hardware}(a)); and (v) an output layer configured for classification or regression.

\subsubsection{Time-Wheel Representation with Packed Two-Generation Slots}
\label{sec:hw_state}
Direct shift-buffer implementations track inter-spike intervals by shifting all buffer entries toward index~0 at each time step, resulting in \textit{\textbf{memory-bandwidth-dominated}} execution. To address this limitation, we adopt a circular time-wheel representation driven by a global pointer increment.
Let $D$ denote the supported delay horizon (e.g., $D=256$ for 8-bit delay precision), and maintain a global pointer $p\in\{0,\dots,D-1\}$ updated once per tick as 
\begin{equation}
    p \leftarrow (p+1)\bmod D.
\end{equation}\label{eq:timestep_index}
In the baseline single-wheel form, each stage is represented by a bit-vector indexed by the wheel position; scheduling sets the bit at 
\begin{equation}
    q=(p+\Delta t)\bmod D, 
\end{equation}\label{eq:sched_index}
while ``due'' is tested at the current pointer position.

\paragraph{Packed two-generation}
When the sample length $L\gg D$, the pointer wraps multiple times within a sample. To prevent wrap-around ambiguity---i.e., to distinguish expectations due in the current wheel rotation from those scheduled across the wrap boundary---we store two generations of state per slot using \emph{two bits per slot}:
\begin{equation}
S_k[u][x]\in\{0,1\}^2,\qquad k\in\{1,2\},\ x\in\{0,\dots,D-1\}.
\label{eq:packed_state_def}
\end{equation}
Here $k=1$ corresponds to the ``expect spine~1'' stage and $k=2$ to the ``expect spine~2'' stage. The two bits of $S_k[u][x]$ represent membership in two consecutive wheel generations (current and next). 
We maintain a single phase bit $\phi\in\{0,1\}$ indicating which packed bit-plane corresponds to the \emph{current} generation:
\begin{equation}
g \triangleq \phi,\qquad \bar g \triangleq 1-\phi .
\label{eq:phase_def}
\end{equation}

\paragraph{Due condition}
The due-now test, performed at buffer index 0 in the software model, is implemented by reading the packed value at $(u,k,p)$ and testing the current bit-plane:
\begin{equation}
\mathrm{due}(u,k) \iff S_k[u][p]_g = 1.
\label{eq:due_test}
\end{equation}

\paragraph{Scheduling rule}
For a delay $\Delta t\in\{0,\dots,D-1\}$, compute $q=(p+\Delta t)\bmod D$.
The expectation is scheduled into the current or next generation based on overflow of the addition $p+\Delta t$:
\begin{equation}
\begin{cases}
S_k[u][q]_g \leftarrow 1, & \text{if } p+\Delta t < D \quad(\text{no wrap}),\\
S_k[u][q]_{\bar g} \leftarrow 1, & \text{if } p+\Delta t \ge D \quad(\text{wrap}).
\end{cases}
\label{eq:sched_rule}
\end{equation}
This overflow test avoids corner cases that arise when comparing $q$ to $p$ (e.g., at $\Delta t=0$).

\paragraph{Wrap update (generation advance)}
When the pointer wraps ($p=D-1\rightarrow 0$), we toggle the phase $\phi\leftarrow 1-\phi$, thereby exchanging the roles of the current and next packed bit-planes. To preserve the two-generation invariant, the bit-plane being reused for new next-generation schedules is cleared once per wheel rotation; equivalently, after the toggle, the plane indexed by $\bar g$ is reset to zero.

\paragraph{Event-driven memory access}
\label{sec:hw_mem_eff}
Event-driven memory access follows from per-channel adjacency lists: for each spike, the CR streams only the affected $\langle u,s\rangle$ targets, so USM is touched only for units connected to spiking channels.
Memory clearing is amortized over $D$ steps. On wrap, the phase bit $\phi$ is toggled and the plane reused for next-generation scheduling is reset once per wheel rotation (once every $D$ timesteps).

\subsubsection{Hardware Experiments Setup}
The asynchronous hardware prototype of DendroNN was implemented using the GlobalFoundries\textsuperscript{\textregistered} 22FDX FDSOI technology node and evaluated through post-layout simulations with \{QuestaSim\} and \{Innovus\}. Two design configurations were considered: DendroNN (2) and DendroNN (3), corresponding to architectures with two and three spines per unit, respectively. 
Hardware performance metrics were averaged across inference runs on the complete \ac{SHD} test set.
To reduce inference cost, the networks deploy a refractory period. All remaining network hyperparameters can be taken from Table~\ref{tab:model_hparams}.

Reference designs used for comparison include \{Loihi2\}\cite{loihi2}, which employs a digital implementation of synaptic delay--based \acp{SNN}; \{DenRAM\}\cite{denram}, which features \acp{SNN} with analog memristor--capacitor--based dendritic delays; \{ReckOn\}\cite{reckon}, a digital accelerator for spiking recurrent neural networks; and \{ElfCore\}\cite{elfcore}, which implements a digital feed-forward \ac{SNN}.

\textit{\textbf{All hardware metrics presented in Fig.~\ref{fig:dendronn_hardware}(c--e) for \{Loihi2\} and \{DenRAM\} were obtained from their respective publications, where available, and normalized to account for differences in technology node and operating voltage. For comparison, the highest SHD accuracy reported in the literature was adopted. In contrast, hardware synthesis and simulation were performed for \{ReckOn\} and \{ElfCore\}, leveraging their open-source implementations. Hardware metrics and accuracy results were derived from post-layout simulations. During simulation, the on-chip learning functionality of \{ReckOn\} and \{ElfCore\} was disabled.}}

The evaluation metrics include the maximum acceleration factor, defined as the ratio between the processing speed of the computing core and real-time operation, which is 1\,ms per time step for the SHD dataset. The energy per time step represents the average energy consumption required to process a single inference sample. The area cost corresponds to the post-layout core area, excluding pad overhead.

\subsubsection{Hardware Results Analysis}
DendroNN (2) and DendroNN (3) attain test accuracies of 88.1\% and 88.7\% on the SHD dataset, respectively. These performances are on par with delay-based neuromorphic hardware platforms for temporal signal processing, including \{Loihi2\} (88.0\% accuracy) and \{DenRAM\} (87.5\% accuracy). Notably, the results reported for \{DenRAM\} are based on hardware-aware simulations rather than direct hardware evaluations. \{ReckOn\} achieves a slightly lower accuracy of 84.2\%, whereas \{ElfCore\} reaches only 72.8\% test accuracy, reflecting the inherent limitations of feedforward \acp{SNN}.

As shown in Fig.~\ref{fig:dendronn_hardware}(c), DendroNN (3) achieves up to 1.9$\times$ higher throughput than delay-based hardware. This gain is enabled by the time-wheel implementation, which supports fully event-driven memory access. The throughput of DendroNN (2) is slightly lower than that of DendroNN (3) due to its reduced dynamic sparsity. Although \{ElfCore\} attains the highest throughput because of its conventional feedforward architecture, this advantage comes at the expense of substantially lower accuracy.

As illustrated in Fig.~\ref{fig:dendronn_hardware}(d), DendroNN (3) delivers up to 4.1$\times$ higher energy efficiency than delay-based hardware and 2.8$\times$ higher energy efficiency than recurrence-based hardware. Notably, it also outperforms the feedforward-based \{ElfCore\} in terms of energy efficiency. These improvements result from eliminating low--arithmetic-intensity vector operations for per-time-step neuron state updates, which are required in other designs. Instead, the architecture relies exclusively on always-on pointer computations and event-driven memory access, avoiding memory read--modify--write operations or buffer shifting that scale with the time-window length. The superior efficiency of DendroNN (3) is further reinforced by the higher dynamic sparsity of its hidden units, as classifier energy consumption is highly sensitive to spike activity.

Finally, DendroNN achieves over 2$\times$ higher area efficiency than the recurrence-based implementation \{ReckOn\}, since the sequence detection layer does not require storage of quantized weights, neuron membrane potentials, or other complex numerical states. The use of binary weights and binary spike activity substantially reduces area overhead.

\section{Discussions}\label{sec:discussion}
We present a novel system for classification of event-based data, the DendroNN, designed as an abstraction of the sequence detection mechanisms present in biological dendrites.
By identifying spike sequences across differing time scales as spatiotemporal features within the incoming data, the model is able to perform high quality sequence-decoding.
This leads to competitive accuracies across multiple datasets like NeuroMorse, sequential and permuted sequential MNIST, and \ac{SHD} on a much smaller memory footprint when compared to current leading-edge \ac{SNN} architectures, like delay-based or recurrent networks.

Furthermore, the DendroNN to  is developed with neuromorphic hardware in mind.
The model displays high degrees of dynamic and static sparsity and substantial quantization, with the hidden layers being fully binarized and the output layer pruned by 70\% and represented by int8.
Additionally, internal operations of DendroNN units use elemental computing operations only, e.g., AND and scatter-gather operations.

An asynchronous near-memory digital computing architecture based on a time-wheel representation is proposed for DendroNN to fully exploit the model’s inherent advantages. By leveraging fully sparse, event-driven memory access, the architecture achieves over $4\times$ higher energy efficiency than state-of-the-art delay-based hardware on the same task with comparable accuracy, enabled by tight algorithm--hardware co-design and co-optimization.

While in the implementation of the DendroNN unit the sequence defining propwerties cannot be trained through the use of gradients, we deploy a rewiring phase that identifies spike sequences of high informational gain within the training data.
At iso-accuracy, this leads to a drastic reduction in network size compared to when relying on randomly initializing the DendroNN units.
It should be noted that this method of picking sequences, which in essence are the features the network uses to classify the data, relies heavily on the right definition of the selectivity criterion.
We observe stark variations in accuracy across different criteria and parameterizations.
Moreover, it is not obvious how to design an optimal criterion in order to achieve top accuracy during classification after the supervised phase that follows the rewiring and builds on top of the chosen sequences.

The DendroNN is build upon the processing of binary event streams.
When confronted with event-based data with values larger than one, this calls for a respective pre-processing that converts the data into binary events while containing all relevant spatial and temporal information.
This is especially relevant for benchmarks like \ac{SHD} where the originally binary event stream is binned into events of larger values because to facilitate limited time series lengths.
To tackle this problem, we use a slicing method that converts event values into spatial information.
While the resulting performance of DendroNNs on \ac{SHD} shows the success of the method, there remains a minor gap to leading-edge methods like delay-based \acp{SNN}.
We argue that, on top of the limited feature finding via the rewiring phase, this could potentially be caused by loss of information during event value slicing.
Future work should thus also explore angles to mitigate this issue. 

From a hardware perspective, compact and energy-efficient nanoelectronic devices, such as those in \cite{multigate}, are promising for discriminating spatiotemporal pulse patterns. However, the programmability and scalability of such devices must always be carefully considered.

In summary, we achieved the reported co-optimization of accuracy and efficiency by abstracting the sequence detection mechanism found in biological dendrites with the targeted hardware principles in mind.
This demonstrates the huge potential for statically and dynamically sparse algorithms when deployed on dedicated neuromorphic hardware.
Further advancing along this avenue could bring real-world applications ever closer toward extremely low-power and intelligent systems.

\section{Methods}\label{sec:methods}
\subsection{Algorithms}
All code was implemented using the machine learning library Torch.
Experiments were run on single NVIDIA RTX 6000 Ada Generations.

\subsubsection{Model}
\label{sec:methods:model}
Here we will give an in depth overview of the PyTorch implementation of the DendroNN. It consists of two major parts, the sparse binary hidden layer and the DendroNN unit.
The hidden layer connectivity matrix is simply implemented as a feed forward layer with a weight matrix buffer of zeros and ones.
As every unit has $N_S$ number of spines, the weight matrix shape is $(n_\text{units}\cdot N_S, n_\text{in})$.
However, due to the fact that each spine only has a single binary connection, there is only a single non-zero entry per weight matrix buffer line.

The DendroNN unit is implemented as a custom PyTorch module.
To understand the choice of implementation, one has to understand a property central to the processing scheme of those units.
As briefly mentioned in Section~\ref{sec:model}, the ability to process the same sequence multiple times in parallel allows the unit to ignore noise induced and possibly meaningless spikes, drastically improving performance.
For reference, see Figure~\ref{fig:model_hparams}.
The idea is to initialize a binary buffer for every unit of shape $(N_S-1, ||\Delta t||_\infty)$, insert spikes that arrive at a spine $i$ into the respective index $[i+1, \Delta t_i]$, and then shift the entries of that buffer towards lower indices with every timestep.
By doing so, if at any given timestep the buffer has an entry of 1 at any of the spines and at index 0 along the shift dimension, then that means that a spike got accepted at the preceding spine $\Delta t_i$ time steps ago.
Based on that, we can AND incoming spikes at spines of index $i>0$ with the buffer at index $[i, 0]$ and insert the result into the buffer at index $[i+1, \Delta t_{i+1}]$, starting the shifting process for the next spine.
Since the first spike within the sequence is not dependent on any preceding spikes, we can always directly place incoming spikes at the zeroth spine into the buffer, kicking off the sequence detection mechanism.
Finally, the unit's output is given by the ANDing of incoming spikes at the last spine $i=N_S$ and the spike buffer at $[N_S, 0]$.

Crucially, by following this principle of computation, two same sequences that stretch over many timesteps but are only a single timestep apart from eachother can be successfully detected, facilitating the recognition of parallel sequences.
Algorithm~\ref{alg:unit} summarizes the principle once more for the special case of $\Delta T=0$.
As introduced in Section~\ref{sec:model}, it might also make sense to allow the spike to arrive with some uncertainty $\Delta T$ with respect to the expected inter-spike interval.
For this implementation, spike acceptance windows of $\Delta T>0$ are easily implemented by increasing the buffer shape to $(N_S-1, ||\Delta t||_\infty+\Delta T)$ and ANDing the incoming spikes at a spine $i>0$ with $\bigvee_{j=0}^{\Delta T}\text{buffer}[i, j]$.
For this, buffer entries need to be reset since the unit will otherwise be activated $\Delta T$ times.


\begin{algorithm}
\caption{Single DendroNN Unit Forward Call}
\label{alg:unit}
\begin{algorithmic}[1]
\Procedure{Forward}{Input}
    \State Input $\Delta t\gets[\Delta t_0,\ldots]$ \Comment{interspike intervals}
    \State Initialize $t_\text{max}\gets$ Input.shape[0] \Comment{number of timesteps}
    \State Initialize $S \gets$ ZeroTensor(($N_S-1$, $t_\text{max}/N_S-1$)) \Comment{processed spine binary buffer}
    \For{timeslice $x$ in Input} \Comment{x is of shape ($N_S$,)}
        \State $S\gets$ Roll($S$, dim=-1) \Comment{proceed one timestep}
        \State $S$[\ldots,-1] $\gets 0$ \Comment{reset wrapped elements}\\    
        \State Initialize $F\gets x$[0] \Comment{first spine input}
        \State Initialize $L\gets x$[-1] \Comment{last spine input}
        \State Initialize $O\gets x$[1:-1] \Comment{other spines input}
        \State Initialize $E\gets S$[\ldots,0] \Comment{expecting spines}
        \State Initialize $A_F\gets F$ \Comment{accepted input first spine, always accepts}
        \State Initialize $A_O\gets O\land E$ \Comment{accepted input other spines}
        \State Initialize $A\gets$ Concatenate([$A_F$,$A_O$])
        \State $S$[\ldots,$\Delta t$] $\gets A$ \Comment{scatter accepted spines at correct times}
    \EndFor
    \State \Return $L$ \Comment{spikes accepted by last spine $\rightarrow$ \textbf{complete sequence}}
\EndProcedure
\end{algorithmic}
\end{algorithm}

Finally, assuming inter-spike intervals can be represented as a 8 bit integer, we define a networks total memory footprint $M$ as
\begin{equation}\label{eq:memory}
\begin{aligned}
    M &= M_\text{in\_cons} + M_\text{out\_weights} + M_\text{inter-spike\_intervals} + M_\text{DendroNN\_state} + M_\text{integrator\_state}\\
    &= n_\text{units}\cdot(N_S\cdot1\text{bit} + (1 - r_\text{p}) \cdot n_\text{classes}\cdot b_\text{out\_weights} + (N_S - 1)\cdot8\text{bit})\\
    &+n_\text{units}\cdot (N_S-1)\cdot\left\lfloor\frac{L}{\overline{N_S}-1}\right\rfloor\cdot1\text{bit}+n_\text{classes}\cdot8\text{bit},
\end{aligned}
\end{equation}
with $b_\text{out\_weights}$ being the bit width and $r_\text{p}$ the pruning rate of output layer weights and the number of timesteps in the data sample sequence $L$.

\subsubsection{Rewiring Phase}
\label{sec:methods:rewiring_phase}
Given a number of input channels $n_\text{in}$, spines $N_S$, and timesteps $L$, and assuming that each $\Delta t_i$ can at max take on integer values of $\lfloor\frac{L}{N_S-1}\rfloor$, a theoretical expression for the combinatorially possible number of sequences detectable by different DendroNN units is given by 
\begin{equation}
    N_\text{seqs}=N_\text{seqs}^\text{spat}\cdot N_\text{seqs}^\text{temp}=n_\text{in}^{N_S}\cdot\left \lfloor\frac{T_{max}}{N_S-1}\right\rfloor^{N_S-1}.
    \label{eq:num_possible_seqs}
\end{equation}
To find the few interesting among all $N_\text{seqs}$ sequences, the rewiring phase essentially consists of three steps:
\begin{enumerate}
    \item inferring the network with training data,
    \item temporarily freezing sequences that occur often enough,
    \item permanently freezing those that fulfill selectivity criterion.
\end{enumerate}
Step 1 is trivial.
Step 2 involves the introduction of a penalty reward ratio $r_\text{pr}$ as well as a longevity state $L$ for every unit.
The penalty reward ratio defines the ratio of samples the sequence has to occur in order to increase its longevity state.
After a batch of size $B$, the longevity is updated based on the number of occurrences $n_\text{occ}$ of each unit.
\begin{equation}
    L_\text{new} = L_\text{old} + \Delta L = L_\text{old} + n_\text{occ} - r_\text{pr} \cdot (B - n_\text{occ})
\end{equation}
By defining the two thresholds $\vartheta_l$ and $\vartheta_h$, as soon as the longevity of a unit crosses one of them, it is rewired or temporarily frozen, respectively.
Since the occurrence of sequences within the dataset is approximately constant over batches, apart from very small values, we did not find a strong dependency of the threshold values on the resulting model performance.

Step 3 evolves around the analysis of sequence occurrences for different classes.
The motivation here is to set a requirement for the spike count of a unit for a specific class and also in comparison to other classes.
We call this the selectivity of a unit.
Analyzing a unit's selectivity necessitates the definition of a criterion.
First, we calculate the relative occurrence $o$ of a sequence in each class $i$ from the number of emitted spikes in samples of that class $n_S^i$ and the number of samples of that class in the batch $n_i$:
\begin{equation}
    o_i = \frac{n_S^i}{n_i}.
\end{equation}
Finally the occurrence is normalized by dividing by the sum of all occurrences, yielding the unit's selectivity towards the class $S_i$:
\begin{equation}
    S_i = \frac{o_i}{\sum_i o_i}.
\end{equation}
In essence, the goal is to ensure that each class is well represented by a set of sequences, facilitating good classification performances.
Therefore, we search a total of $\frac{n_\text{units}}{n_\text{classes}}$ for every class and omit samples of classes we have already found enough units for during the search for common sequences.
The calculation of selectivities, however, is based on data samples of all classes.

Based on a set of selectivities $\{S_0, S_1, \ldots\}$, we can then define a criterion that a unit has to fulfill in order to be permanently frozen.
We primarily tested two criteria:
\begin{enumerate}
    \item The sequence is selective towards a class if its class-dependent selectivity is greater than a pre-defined selectivity threshold $\vartheta_S$:
    \begin{equation}\label{eq:criterion_std}
        S_i>\vartheta_S.
    \end{equation}
    \item In addition to $S_i$ being greater than $\vartheta_S$, it also has to have a positive gap $\epsilon_S$ towards the selectivity of that unit towards all other classes: 
    \begin{equation}
        S_i>\vartheta_S\land S_i-\epsilon_S>||\tilde{S}||_\infty,
    \end{equation}
    with $\tilde{S}=\{S_0, \ldots, S_{n_\text{classes}}\}\setminus\{S_i\}$.
\end{enumerate}
Although criterion 2 promises much more discriminative spatiotemporal features, we found the classification performance to not deteriorate using criterion 1 in addition to the rewiring phase terminating much quicker.
Therefore, for all our benchmarking runs, we applied criterion 1.
Note that, due to the overall lower selectivities necessary to fulfill criterion 2, it might offer the more efficient option in terms of number of emitted events.

\begin{figure}
    \centering
    \begin{tikzpicture}
\begin{groupplot}[
    group style={
        group size=3 by 1,
        horizontal sep=1.6cm,
        vertical sep=1.0cm,
    },
    width=5cm,
    height=5cm,
    enlargelimits=false,
    ymin=0.89,
    ymax=1.02,
    every axis y label/.style={at={(axis description cs:-0.29,0.5)}, rotate=90, anchor=center},
    grid=none,
    every axis x label/.style={at={(axis description cs:0.5,-0.2)}, rotate=0, anchor=center},
    grid=none,
    every axis plot/.append style={line width=1pt},
    no markers,
    ylabel={norm. accuracy},
]

\nextgroupplot[xlabel={$\vartheta_S$}, xtick={0.05, 0.08, 0.11}, xticklabels={0.05, 0.08, 0.11}]
\addplot+[solid, black] coordinates {(0.03, 0.9423) (0.05, 0.9468) (0.07, 0.9852) (0.09, 1.0) (0.11, 0.9818) (0.13, 1.0)};

\nextgroupplot[xlabel={$r_\text{pr}$}, xmode=log]
\addplot+[solid, black] coordinates {(16, 0.9276) (8, 0.9071) (4, 0.9302) (2, 0.9523) (1.0, 0.9872) (0.5, 0.9677) (0.25, 0.9933) (0.125, 1.0)};    

\nextgroupplot[xlabel={$r_\text{pr}$}, ylabel={$\langle S_\text{max} \rangle$}, xmode=log, ymin=0.03, ymax=0.3,]
\addplot+[solid, black] coordinates {(16, 0.0531) (8, 0.1244) (4, 0.1545) (2, 0.1811) (1.0, 0.1979) (0.5, 0.2171) (0.25, 0.2759) (0.125, 0.2781)};    

\end{groupplot}
\end{tikzpicture}
    \caption{Visualization of the impact of different rewiring phase parameterizations on the model's classification performance. $\langle S_\text{max} \rangle$ is the mean of maximum observed class selectivities during the rewiring.}
    \label{fig:rewiring_hparams}
\end{figure}
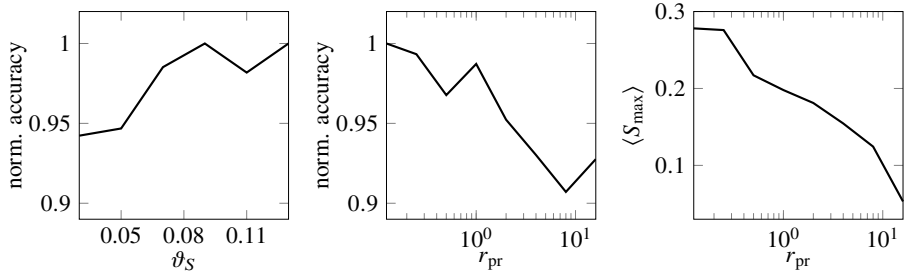

Figure~\ref{fig:rewiring_hparams} visualizes the influence of $\vartheta_S$ and $r_\text{pr}$ on the accuracy of the network resulting from the rewiring phase.
As visible, both hyperparameters have strong impact on the resulting performance.
It should be noted, that, in order to obtain the data for $\vartheta_S$, we enforced units to have a selectivity within $[\vartheta_S,\vartheta_S+0.02]$.
Normally, criteria are imposed and the most selective units are chosen.
However, $\vartheta_S$ has little to none influence on the largest observed selectivity leading to a rather flat curve.
Therefore, this graph rather reflects the importance of enforcing selective units than the actual influence of parameterizing correctly.
$p_\text{pr}$ seems to increase with smaller values before flattening out, hinting at the fact that too large penalties negatively impact the quality of the resulting sequences as spatiotemporal features.
This is underlined the furthermost right graph, showing the relationship between $r_\text{pr}$ and the resulting mean maximum observed class selectivities: units that are penalized stronger making them more present across different samples tend to be specialized worse toward specific classes.
Due to this trend, we set $r_\text{pr}$ to 0.1 for all trainings.
Finally, while the influence on accuracy is obvious, one should note that the choice of parameterization can also have a large influence on the total time needed for rewiring.

Whenever a unit is discarded and rewired, its binary connections and inter-spike intervals are randomly drawn from a uniform distribution.
Updating the upper bound the inter-spike intervals are drawn from according to what classes are still being inferred and what sequence length makes sense can heavily decrease search time leading to an overall quicker termination of the phase.
If the number of spines is also within a range of values $[N_S^\text{min}, N_S^\text{max}]$ rather than being constant, it is drawn from a multinomial distribution with probabilities $p_i=0.3^{N_S^i}/\sum_j0.3^{N_S^j}$.
A good example of where the number of spines should be drawn from a distribution rather than being constant is the following:
Interestingly, the shortest words included in the NeuroMorse dataset, being "a" and "I", consist of only two symbols.
This necessitates a minimal number of spines of two for networks trained on NeuroMorse.
Since sequences of two spikes are not very discriminative, we hence set the maximal number of spines to five.
This results in the spine count of rewired units being drawn from a Multinomial distribution distribution between between two and five.

\subsubsection{Supervised Training and Finetuning}
The output layer consists of a feed-forward linear layer connected to a layer of integrator units of size corresponding to the number of classes in the dataset.
The integrator units transform the network's temporal output into a static class prediction.
Their output at the last time step of inference is taken as the prediction and used for categorical cross entropy loss calculation.
Contrasting the DendroNN units, the output weights can be trained using backpropagation through time.
All networks are trained with a learning rate of $10^{-3}$; utilizing a scheduler does not yield any significant advantage.
We found that AdamW provided slightly better results than Adam.
Further, the use of dropout did not result in any performance increase.

For compression, the output layer is pruned iteratively, with repeating pruning steps of 5\% and a single finetuning epoch per step.
Finally, quantization to a 8\,Bit integer representation happens post-training.
We also tried quantization aware training, but documented better results from post-training quantization.

\subsubsection{Model and Rewiring Hyperparameters}
\label{sec:methods:hparams}
Table~\ref{tab:model_hparams} provides an overview of the hyperparameters chosen for the model architecture and rewiring phase for all datasets.

\begin{table}[h]
    \centering
    \caption{Overview of the model hyperparameters chosen for the results of experiments presented in this work.}

\begin{tabular}{l c c c}
\toprule
\textbf{Hyperparameter} & \textbf{NeuroMorse} & \textbf{sMNIST \& p-sMINST} & \textbf{SHD} \\
\midrule
\#Units & 200 & 1500 & 3000 \\
max. \#Spines & 5 & 5 & 2 \\
min. \#Spines & 2 & 5 & 3 \\
$\Delta T$ & 2 & 0 & 0 \\
$r_\text{pr}$ & 0.1 & 0.1 & 0.1 \\
criterion & \eqref{eq:criterion_std} & \eqref{eq:criterion_std} & \eqref{eq:criterion_std} \\
$\vartheta_S$ & 0.025 & 0.1 & 0.05 \\
$r_\text{p}$ & 0.0 & 0.7 & 0.7 \\
$b_\text{out\_weights}$ & 32 & 8 & 8 \\
\bottomrule

\end{tabular}

    \label{tab:model_hparams}
\end{table}




\subsubsection{Handling Non-Binary Data}\label{sec:methods:slicing}
Some data might contain events which carry values larger than one.
Since DendroNNs only process binary event streams, this calls for a pre-processing scheme that converts the data into binary event streams while retaining the information of the original event values for high quality classification.
For this, we introduce a technique which we call slicing.
In principle, slicing can be applied to all non-binary data..

To apply slicing to any data tensor, we define a set of lower and higher slice borders $\{a_1, a_2, ...\}$ and $\{b_1, b_2, ...\}$, respectively.
The data is then duplicated along the input channel axis times the number of slices.
In each slice $i$, events for whose value $x\in(a_1,b_2)$ is true are retained in that slice as a binary event, otherwise the tensor entry is set to 0.
After applying this pre-processing scheme, the input data shape has increased by the number of slices, with binary events corresponding to ever larger original event values as you go up along the input channel axis.
Pseudo code of the slicing scheme is provided in Algorithm~\ref{alg:slicing}.
Although there exist many ways to encode multi-valued input data into binary spikes, we found this method to yield the highest accuracies.
For example, spike encoding the data through a single layer of LIF neurons led to a significant drop in performance compared to slicing. Applying this technique also has a huge impact on the possible selectivities of found units: without slicing, just binarizing all events, found sequences display no discriminatory ability leading to random-chance accuracies.


\begin{algorithm}
\caption{Input Data Slicing}
\label{alg:slicing}
\begin{algorithmic}[1]
\Procedure{Slicing}{Input}
    \State Input $A\gets[a_1, a_2, \ldots]$  \Comment{lower slice boarders}
    \State Input $B\gets[b_1, b_2, \ldots]$  \Comment{upper slice boarders}
    \State Initialize $N\gets$ len($A$)  \Comment{number of slices}
    \State Initialize $S\gets$ Input.shape
    \State Initialize $C\gets S$[-1]  \Comment{number of input channels}
    \State Initialize $D\gets$ ZeroTensor(($S$[0], $S$[1],\ldots,N*C))
    \For{$i = 0$ to $N$}
        \State Initialize $L\gets\text{Input}>A$[$i$]  \Comment{lower thresholding}
        \State Initialize $U\gets\text{Input}<B$[$i$]  \Comment{upper thresholding}
        \State $D$[\ldots, $i$*$C$:$(i+1)$*$C$] $\gets L\land U$  \Comment{create slice $i$}
    \EndFor
    \State \Return $D$
\EndProcedure
\end{algorithmic}
\end{algorithm}

\subsubsection{Data Preparation}\label{sec:methods:data}
\paragraph{NeuroMorse}\label{sec:methods:data:neuromorse}
The data is prepared according to the original publication and the corresponding data repository \cite{neuromorse}.
The code translates morse code so that each dot and dash contribute a single spike in input channel 0 and 1, respectively.
Inter-symbol intervals are five timesteps while inter-letter intervals are ten timesteps.
The parameterization of all three noise types and levels is also taken from the original NeuroMorse repository.
Samples are zero padded to match the maximal possible length words included in the dataset.
For testing on the noisy dataset variants, the network was, however, still trained on noise-free data.
Finally, the dataset is prepared with a batch size of 50.

\paragraph{sMNIST and p-sMNIST}
\label{sec:methods:data:mnist}
The sequential variants of MNIST are prepared by simply loading the original MNIST dataset and flattening each tensor into a 1D tensor of length 784, representing the time series.
We apply no data augmentation.

Notably, the \ac{smnist} variants are commonly prepared with multiple input channels, so that each input channel activates for specific greyscale values of the input \cite{adaptive_srnn}.
As the number of input channels quickly reaches high double digits, this leads to larger input layers and an overall increased memory footprint of the networks.
As written in Section~\ref{sec:experiments:mnist}, pixel values larger than 0 are quantized to 1 to facilitate the binary event processing of the DendroNN model.
This precoessing scheme presents a special case of slicing as described in Section~\ref{sec:methods:slicing}.
For DendroNNs, this means that the samples are truly processed through a single input channel. 
One could raise the argument that, similar to the \ac{SHD} data, information is caught within the greyscale value and gets lost completely due to the binary quantization of input data.
We argue that, while for \ac{SHD} the pixel value indicates the intensity of the incoming audio signal which is central for a correct classification, for MNIST all pixel values between 1 and 255 indicate the presence of the number to be detected, allowing for such an aggressive quantization without significant performance degradation.
Finally, we prepare the datasets in batches of 256.

\paragraph{SHD}
\label{sec:methods:data:shd}
The \ac{SHD} samples are loaded using the Tonic framework~\cite{tonic}.
The spikes of each sample are binned to 8\,ms frames.
Furthermore, the input channels are downsampled from 700 to 100 channels.
Resulting samples of different sequence lengths are zero padded to a pre-defined maximal length.
To reduce the triggering of units by noise-induced spikes, we apply Tonic's denoising transform.
Apart from that, we apply no further augmentation.
All batches are prepared with a size of 256.

Similarly to the \ac{smnist} variants and as already introduced in Section~\ref{sec:experiments:shd}, the \ac{SHD} samples usually carry events of values much larger than 1.
For this reason we apply slicing with one slice interval per integer value contained in the dataset.
Given a slice center $c$, we create lower and upper slice boarders by subtracting and adding a base value of 3 to scaled slice centers of $c/1.3$ and $c\cdot1.3$, respectively.
This creates a rather large overlap of individual slices.
However, we found that this leads to the best results, possibly due to robustness against event value fluctuation across different data samples.

\subsubsection{Null Class Detection in NeuroMorse}\label{sec:methods:nullclass}
When standard \acp{SNN} face input of a class they have never seen before during training they produce a reasonable output vector predicting that the class is one that was included during training.
This assumes that the density of input spikes is roughly the one of learned samples.
DendroNNs, on the other hand, look for spatiotemporal spike sequences they have learned during the rewiring phase.
If they receive input of a null class sample, most of those spike sequences will be absent, leaving the network output rather silent.
Interestingly, this absence of output basically presents a measure of the network's uncertainty.
We utilize this by defining an output threshold for each class and defining that if the output of all classes stay below those thresholds, the input is classified as a null class sample.
As seen by the good classification accuracies, this technique is rather successful.
We also made sure that the accuracy across all noise levels is not purely dominated by null class predictions, as, as written above, the network could score decently by turning to only predicting null classes after a certain level of noise.
Instead, the prediction is always well balanced between learned and null classes.
Based on the interaction with that mechanism it is, however, hard to argue that DendroNNs can deal with noise any better than \acp{SNN}.
\subsection{Hardware}

\subsubsection{Computation Model}
\label{sec:hw_model}
As defined in Sections~\ref{sec:model} and~\ref{sec:methods:model}, each unit maintains a binary temporal buffer of size $(N_S-1,\lVert\Delta t\rVert_\infty)$. Spikes arriving on spine $i>0$ are gated by the \emph{due-now} entry of stage $i$; upon a successful match, the partial sequence advances by writing an expectation into the subsequent stage at an offset determined by $\Delta t_i$. An output spike is generated when an input on the final spine coincides with the \emph{due-now} entry of the last stage. For the special case $N_S=3$, each unit maintains two $(N_S-1)$ state stages, corresponding to expectations for spine~1 and spine~2. When the spike acceptance window is set to $\Delta T=0$, matching is exact and requires no windowed reduction, as the coincidence test involves only a single \emph{due-now} bit.

\subsubsection{Memory Organization}
\label{sec:hw_mem}
In CR SRAM, sparsity is exploited by representing connectivity with per-channel adjacency lists. The array \texttt{chan\_ptr} defines contiguous segments within a compact list \texttt{conn\_list} of pairs $\langle u, s\rangle$, where each pair indicates that channel $c$ projects to spine $s \in \{0,1,2\}$ of unit $u$. Upon observing an event $(t, c)$, the CR streams only the corresponding $\langle u, s\rangle$ targets to the UE, thereby avoiding any scan over non-activated units. The USM stores $S_1$ and $S_2$ in on-chip SRAM using bit packing; each slot holds two bits, one per generation, and updates are applied via masked read--modify--write (RMW) operations on wide words.

\subsubsection{Working Flow and Datapath}
\label{sec:hw_flow}
At each time-bin $t$, the system advances the pointer $p$ and processes all events in that bin. For each input event $(t,c)$, the CR retrieves the adjacency range $[\texttt{ptr}[c],\texttt{ptr}[c+1])$ and streams the corresponding $\langle u,s\rangle$ pairs to the UE. Each streamed target triggers one of three micro-operations determined by the spine index $s$. Let $g=\phi$ denote the current generation index.

\paragraph{Spine 0 ($s=0$): schedule expectations for stage $k=1$.}
Spine~0 initiates candidate sequences by scheduling an expectation for spine~1. The UE reads $\Delta t_0[u]$, computes $q$ via~\eqref{eq:sched_index}, and applies~\eqref{eq:sched_rule} to stage $k=1$:
\begin{equation}
\begin{cases}
S_1[u][q]_g \leftarrow 1, & p+\Delta t_0[u] < D,\\
S_1[u][q]_{\bar g} \leftarrow 1, & p+\Delta t_0[u] \ge D.
\end{cases}
\label{eq:spine0_op}
\end{equation}

\paragraph{Spine 1 ($s=1$): due-check, consume, schedule stage $k=2$.}
To realize coincidence gating, the UE reads the due bit
\begin{equation}
d \leftarrow S_1[u][p]_g .
\label{eq:spine1_due}
\end{equation}
If $d=1$, it consumes the partial match by clearing $S_1[u][p]_g\leftarrow 0$, then reads $\Delta t_1[u]$, computes $q$ via~\eqref{eq:sched_index}, and schedules stage $k=2$:
\begin{equation}
\begin{cases}
S_2[u][q]_g \leftarrow 1, & p+\Delta t_1[u] < D,\\
S_2[u][q]_{\bar g} \leftarrow 1, & p+\Delta t_1[u] \ge D.
\end{cases}
\label{eq:spine1_op}
\end{equation}
If $d=0$, the event has no effect.

\paragraph{Spine 2 ($s=2$): due-check, consume, emit unit spike.}
The UE reads
\begin{equation}
d \leftarrow S_2[u][p]_g .
\label{eq:spine2_due}
\end{equation}
If $d=1$, it clears $S_2[u][p]_g\leftarrow 0$ and emits a unit spike for unit $u$; otherwise, the event has no effect.

This packed representation preserves parallel sequence processing: multiple in-flight partial matches correspond to multiple future slots marked in $S_1$ and $S_2$ (potentially across both generations) without additional per-unit queues.

\subsubsection{Hardware Baselines}
\label{baseline}
State-of-the-art hardware platforms for efficient long-term memory in temporal signal processing include the following devices and systems:

\textbf{Multi-gate FeFET}~\cite{multigate}: This dendrite-inspired device discriminates spatiotemporal pulse patterns and supports parallel processing in three-dimensional neuromorphic architectures. It uses a segmented multi-gate FeFET with a ferroelectric layer to detect consecutive input pulse sequences. Although experiments show effective emulation of selective sequence discrimination seen in cortical neuron dendrites, system-level evaluations on the target task were not reported and are therefore outside the scope of this comparison.

\textbf{Loihi2}~\cite{loihi2}: An end-to-end framework that enables event-driven training of spiking neural networks with synaptic delays on Intel’s Loihi2 neuromorphic processor~\cite{loihi2_hw}. It represents the state of the art in \textbf{digital} hardware architectures supporting \textbf{synaptic delays}.

\textbf{DenRAM}~\cite{denram}: An analog feed-forward SNN implemented in 130\,nm technology, integrating dendritic compartments with resistive RAM (RRAM) to realize both synaptic weights and delays. It represents the state of the art in \textbf{analog} hardware approaches based on memristive \textbf{delayed synapses}.

\textbf{ReckOn}~\cite{reckon}: A digital spiking recurrent neural network processor implemented in 28\,nm technology that supports task-agnostic online learning over multi-second timescales. It represents the state of the art in \textbf{spiking RNN} hardware implementations.

\textbf{ElfCore}~\cite{elfcore}: A feed-forward SNN processor implemented in 28\,nm technology, featuring a local self-supervised learning engine for multilayer temporal learning without labeled data. High arithmetic intensity and energy efficiency are achieved through operator fusion, making it representative of the state of the art in \textbf{spiking feed-forward} hardware implementations.

\subsubsection{Technology and Voltage Normalization}
\label{subsec:normalization}

To enable fair comparison across hardware platforms implemented in different
technology nodes and supply voltages, reported metrics were normalized to a
common reference using first-order CMOS scaling laws. Dynamic energy was
assumed to scale as $E \propto CV^2$, with capacitance proportional to feature
size.

Energy metrics were normalized as
\begin{equation}
E_{\mathrm{norm}} = E_{\mathrm{orig}}
\cdot \frac{\mathrm{node}_{\mathrm{ref}}}{\mathrm{node}_{\mathrm{orig}}}
\cdot \left(\frac{V_{\mathrm{ref}}}{V_{\mathrm{orig}}}\right)^2 .
\end{equation}

Power was normalized using the same scaling under constant frequency
assumption, while area was scaled quadratically with technology node:
\begin{equation}
A_{\mathrm{norm}} = A_{\mathrm{orig}}
\cdot \left(\frac{\mathrm{node}_{\mathrm{ref}}}{\mathrm{node}_{\mathrm{orig}}}\right)^2 .
\end{equation}

Latency was not normalized due to design-dependent voltage--delay behavior.

\subsubsection{Parameter Scalability vs. Real Silicon Cost}
\label{subsec:silicon efficiency}

Table~\ref{tab:results_SHD} highlights the significant improvement in model scalability as reflected by the reduction in parameter count.
Figure~\ref{fig:dendronn_hardware}(e) further illustrates the tape-out-ready, post-layout silicon area of the DendroNN computing core, demonstrating its hardware-level area efficiency under equivalent—indeed enhanced—functionality.
Nevertheless, improvements in parameter efficiency do not directly translate into proportional reductions in actual silicon area.
This discrepancy stems from the fact that parameter memory alone does not fully represent the true silicon cost. 
The overall hardware footprint also encompasses the MAC array, intermediate result buffers, neuron state memory, and additional supporting logic. 
Accordingly, we report the complete post-layout silicon area to provide a comprehensive and realistic evaluation of hardware cost.

\section*{Acknowledgments}\label{sec:acknowledgments}
Jann Krausse 
is funded in the frame of the Important Project of Common European Interest on Microelectronics and Communication Technologies (IPCEI ME/CT). The IPCEI ME/CT is funded by the German Federal Ministry for Economic Affairs and Energy, Regional Development and Energy, the Ministry of Economic Affairs, Industry, the Saxon State Ministry for Economic Affairs, Labour, Energy and Climate Action and the European Union within “NextGenerationEU”.
Kyrus Mama was funded through a Stanford Graduate Fellowship.
We acknowledge the CapoCaccia Workshop Towards Neuromorphic Intelligence 2025 for providing a wonderful platform to connect and develop some of the ideas that led to this work.
Furthermore, we thank the FZI Research Center for Information Technology for providing the compute infrastructure used for neural network trainings.
We also thank Kwabena Boahen for reviewing the manuscript and Moritz Neher for fruitful discussion during algorithm development.









\printbibliography


\end{document}